\documentclass[runningheads]{llncs}

\usepackage{eccv}

\usepackage{eccvabbrv}

\usepackage{graphicx}
\usepackage{booktabs}
\usepackage{multirow}
\usepackage{colortbl}
\usepackage{wrapfig}
\usepackage{caption}

\usepackage[accsupp]{axessibility}  

\usepackage[pagebackref,breaklinks,colorlinks,citecolor=eccvblue]{hyperref}

\usepackage{orcidlink}
\definecolor{lightgray}{gray}{0.9}

\begin{document}

\title{CityGuessr: City-Level Video Geo-Localization \\on a Global Scale} 

\titlerunning{CityGuessr}

\author{Parth Parag Kulkarni\inst{1} \and
Gaurav Kumar Nayak\inst{2} \and
Mubarak Shah\inst{1}}

\authorrunning{Kulkarni et al.}

\institute{Center for Research in Computer Vision, University of Central Florida, USA \and
Mehta Family School of DS \& AI, Indian Institute of Technology Roorkee, India
\email{parthparag.kulkarni@ucf.edu; gauravkumar.nayak@mfs.iitr.ac.in; shah@crcv.ucf.edu}}

\maketitle

\vspace{-0.3in}
\begin{abstract}
  Video geolocalization is a crucial problem in current times.  Given just a video, ascertaining where it was captured  from can have a plethora of advantages. The problem of worldwide geolocalization has been tackled before, but only using the image modality. Its video counterpart remains relatively unexplored. Meanwhile, video geolocalization has also garnered some attention in the recent past, but the existing methods are all restricted to specific regions. This motivates us to explore the problem of video geolocalization at a global scale. Hence, we propose a novel problem of worldwide video geolocalization with the objective of hierarchically predicting the correct city, state/province, country, and continent, given a video. However, no large scale video datasets that have extensive worldwide coverage exist, to train models for solving this problem. To this end, we introduce a new dataset, ``\textit{CityGuessr68k}'' comprising of 68,269 videos from 166 cities all over the world. We also propose a novel baseline approach to this problem, by designing a transformer-based architecture comprising of an elegant ``\textit{Self-Cross Attention}'' module for incorporating scenes as well as a ``\textit{TextLabel Alignment}'' strategy for distilling knowledge from textlabels in feature space. To further enhance our location prediction, we also utilize soft-scene labels. Finally we demonstrate the performance of our method on our new dataset as well as Mapillary(MSLS)\cite{warburg2020mapillary}. Our code and datasets are available \href{https://github.com/ParthPK/CityGuessr}{here}.
  \vspace{-0.1in}
 \keywords{CityGuessr \and geolocalization \and Self-Cross Attention \and soft-scene labels \and TextLabel Alignment}
\end{abstract}

\vspace{-0.32in}
\section{Introduction}
\label{sec:intro}
\vspace{-0.1in}
Geolocalization refers to the process of determining the geographic position of a sample, which can be an image, a video or a text description of a place. If the input sample is strictly visual, i.e., an image or a video, the problem is also termed as Visual Place Recognition (VPR). Geolocalizing images has gained popularity over time, witnessing substantial advancements in the field. In contrast, video geolocalization is currently in its early stages of development. The relevance of video geolocalization today cannot be understated. When the origin of a video is unknown, determining the part of the world the video was recorded in, can assist in a variety of investigative and exploratory applications. The current transformation of social media has resulted in an explosion of video content, making it a valuable resource. Videos also tend to have more visual information as compared to images, owing to the temporal context that images lack. This makes the problem of video geolocalization even more essential. 

There are different levels of granularities in geolocalization problems, right from street identification to worldwide geolocalization, each having its own significance. Image geolocalization has been attempted on both ends of the spectrum\cite{liu2019lending}\cite{shi2020looking}\cite{yang2021cross}\cite{zhu2022transgeo}\cite{zhu2023r2former}\cite{muller2018geolocation}\cite{pramanick2022world}\cite{clark2023we}, with varying levels of success using different approaches catering to the specific problems at hand. Same isn't the case with video geolocalization. There has been some research at the fine-grained level\cite{regmi2021video}\cite{vyas2022gama}\cite{zhang2023cross}, but the same problem at the global level remains largely \textit{unsolved}. Thus, in this work we formulate a unique problem of \textit{worldwide video geolocalization} in an attempt to leverage the information affluence of video domain to address this issue. 

Geolocalization in general can be performed in two ways. Retrieval is the more popular approach where a query input is compared with a gallery of known references, which gives the location of the query, provided we are able to find the best match. Although retrieval approaches are more accurate at the fine-grained level, they tend to be computationally expensive and depend heavily on the domains of the queries and references. Any domain shifts tend to have massive repercussions which can snowball quickly into larger issues. Also,  constructing a gallery that covers the entire world is not feasible. The second approach, i.e., classification overcomes these limitations. Classification constitutes dividing the region of interest, which in our case is the entire world, into classes, be it in the form of places, or literal partitions of the globe; and identifying the class a sample belongs to. Geolocalization via classification not only decreases compute, but also covers the entire world with ease. As an added advantage, classification can be performed at different hierarchies (city level, state/province level, country level, continent level, etc.), enabling the user to adjust as per their requirements. 

Many recent works focusing on the problem of image-geolocalization\cite{weyand2016planet}\cite{seo2018cplanet}\cite{muller2018geolocation}\cite{pramanick2022world}\cite{clark2023we}, have proposed classification-based methods for this reason.  A general classification pipeline includes an encoder backbone (CNN\cite{NIPS1989_53c3bce6} or transformer\cite{dosovitskiy2020image}-based) to obtain a feature embedding of the input image, and an MLP\cite{kawaguchi2000multithreaded} for class prediction. Previous works also have some additional components or changes in the architecture to aid the classification task, like incorporation of scenes. Keeping this in mind we propose a very unique way of incorporating scenes in our model. Text on the other hand, is relatively unexplored for aiding geolocalization. Intuitively, humans are more likely to identify a location, if they can associate a name to a picture/video of that place which they have previously seen. Consequently, it follows that distilling knowledge from text into a geolocalization model would enhance the model's prediction capability. This motivates us to incorporate text from labels, i.e. city/state/country/continent names during the training procedure by aligning the features of our model to the text embedding of labels without the use of any additional information. 

In this paper, we propose a classification-based approach for video geolocalization at global scale. Our objective is to predict the city in which an input video was recorded, and subsequently, the above mentioned hierarchies of state/province, country and continent. Our proposed method comprises of a transformer\cite{dosovitskiy2020image}-based model, with a novel Self-Cross Attention module for scene prediction to assist with the training. We also incorporate soft-labels for computing scene prediction loss during training of our model. A TextLabel Alignment strategy is also implemented for feature enhancement. To the best of our knowledge this is the first attempt to solve this problem, and hence it can serve as a baseline approach for future research. 

An obstacle in solving this problem, is the absence of a 
large scale worldwide dataset for video geolocalization. Existing video geolocalization datasets focus only on specific regions, like BDD\cite{yu2020bdd100k} in California and New York, USA; KITTI\cite{geiger2012we} in  Karlsruhe, Germany; Brno-Toyota\cite{ligocki2020brno} in Brno Czech Republic; and Seqgeo\cite{zhang2023cross} in Vermont, USA. This dense coverage of a limited area works well for retrieval-based geolocalization approaches, but it restricts the data domain to limited parts of the world. To train any model on the global scale, geographic coverage is essential as it exposes the model to a diverse set of locations with variations in environment, infrastructure and salient features instrumental in generalization of the approach. In that context, Mapillary(MSLS)\cite{warburg2020mapillary}, is an image sequence dataset, which covers $30$ cities around the world. The geographical coverage, although more than the previously mentioned datasets, is still lacking. The number of sequences in Mapillary is also relatively small, which limits its scope for large scale generalized training. Thus there is a requirement for a large scale global level dataset with a substantial geographical coverage. To this end, we propose CityGuessr$68$k consisting of $68,269$ videos from $166$ cities all over the world. We also provide soft-scene labels for all video samples in the dataset.

Our main contributions can be summarized as follows:
\begin{itemize}
    \item We formulate a 
    novel problem of worldwide video geolocalization
    \item 
    To benchmark this new problem, we introduce the first global-scale video dataset named `CityGuessr$68$k', containing 
    $68,269$ videos from $166$ cities.
    \item We propose a 
    baseline approach with a transformer-based architecture 
    with two primary components 
    \begin{itemize}
        \item \textit{Self-Cross Attention} module for incorporating scenes (which leverages soft-scene labels for location prediction enhancement)
        \item \textit{TextLabel Alignment} strategy for distilling knowledge from textlabels in feature space
    \end{itemize}
    \item We demonstrate the efficacy of our model with performance results on \\ CityGuessr68k as well as Mapillary(MSLS) datasets.
\end{itemize}
\vspace{-0.25in}
\section{Related Work}
\label{sec:relw}
\vspace{-0.1in}
Geolocalization can be approached by classification or retrieval. Existing video geolocalization methods are retrieval-based and focus on fine-grained localization. Whereas, all worldwide geolocalization methods are classification-based and cater exclusively to images. We will briefly discuss the relevant works below. 
\vspace{-0.2in}
\subsubsection
{Video Geolocalization} It is a young field with very few works, all of which are retrieval-based. Retrieval approaches can either be same-view or cross-view, depending on the domain of the query and reference images. Cross-view has been more popular due to ease of obtaining a reference gallery of satellite images, as compared to ground-view images. Earlier cross-view image models\cite{workman2015wide}\cite{liu2019lending}\cite{hu2018cvm}\cite{regmi2019cross}\cite{regmi2019bridging}\cite{toker2021coming} \cite{shi2020looking} were CNN-based models, while introduction of ViT\cite{dosovitskiy2020image} gave rise to new transformer-based approaches\cite{yang2021cross}\cite{zhu2022transgeo}. 
Although Gonzalo et. al.\cite{vaca2012city} had proposed a solution for trajectory prediction of a moving camera using Bayesian tracking and Minimum Spanning Trees, until recently, video geolocalization using deep learning was relatively unexplored. GTFL\cite{regmi2021video} was such a video geolocalization model, which used a hybrid architecture based on VGG\cite{simonyan2014very} along with self-attention, to solve frame-to-frame same-view video geolocalization. Gama-Net\cite{vyas2022gama} extended it to frame-to-frame cross-view video settings by using a hybrid(ResNet\cite{he2016deep}/R3D\cite{6165309}/ViT\cite{dosovitskiy2020image} based) network. Both these methods use BDD100k\cite{yu2020bdd100k} as a source for query videos, while obtaining reference gallery from other sources. Recently, Seqgeo\cite{zhang2023cross} proposed a clip-to-area model instead of a frame-to-frame approach. However, all these video geolocalization techniques are fine-grained being limited to very few locations. Scaling of these methods to global level is very difficult due to high computation costs and infeasible requirement of a very large reference gallery that covers the whole world.\\
\vspace{-0.3in}
\subsubsection{Worldwide geolocalization} It has been image-exclusive since its inception. Weyand et. al\cite{weyand2016planet} first introduced a classification-based approach on the Im2GPS\cite{hays2008im2gps} dataset. Vo et. al\cite{vo2017revisiting} introduced classification in multiple hierarchies, while on the other hand CPlaNet\cite{seo2018cplanet} introduced a combinatorial partitioning technique for combining coarse hierarchies to predict finer ones. Till this point, visual input was the exclusive information available to the model to perform classification. ISNs\cite{muller2018geolocation} incorporated scenes, by using three separate encoders for each corresponding scene, depending on whether the image was `indoor', `natural' or `urban'. The concept of hierarchical evaluation, i.e., using coarser hierarchies to refine finer predictions, was also introduced in this paper. Translocator\cite{pramanick2022world} used an additional input of segmentation maps along with images, training twin encoders. Recently, GeoDecoder\cite{clark2023we} introduced a completely novel encoder-decoder architecture, which incorporates scenes and hierarchies as queries to the decoder which are attended to by the encoded image features to give seperate embeddings for individual queries, for scene prediction as well as geolocalization.

All recent works attempting to solve this problem train their models on MP-16\cite{larson2017benchmarking}(except for \cite{seo2018cplanet}), while datasets like Im2GPS\cite{hays2008im2gps}, Im2GPS3k and YFCC4k\cite{vo2017revisiting}, and YFCC26k\cite{theiner2022interpretable} are popular for validation. Clark et. al\cite{clark2023we} proposed a new validation dataset, GWS15k which is a more balanced dataset with more worldwide coverage. Note that there are no video datasets in this domain. Video datasets like BDD\cite{yu2020bdd100k}, KITTI\cite{geiger2012we}, Brno-Toyota\cite{ligocki2020brno} and Seqgeo\cite{zhang2023cross} capture driving and/or walking videos, but all of them are restricted to just one or two geographical regions which make them relevant for training region-scale geolocalization models, but they cannot be used for training geolocalization models at a global scale. While Mapillary(MSLS)\cite{warburg2020mapillary} dataset is based in multiple cities around the world, it is an image sequence dataset(We do repurpose MSLS to suit our problem and perform experiments on it as described in Section \ref{sec:map}). We aim to mitigate this issue, by proposing a global scale dataset which extensively covers different regions of the world, which we call CityGuessr68k. Our dataset can be used as the primary benchmarking dataset for this novel task. 
\vspace{-0.2in}
\section{CityGuessr68k Dataset}
\label{sec:cg68}
\vspace{-0.1in}
\begin{figure}[t]
  \centering
  \includegraphics[width = \textwidth]{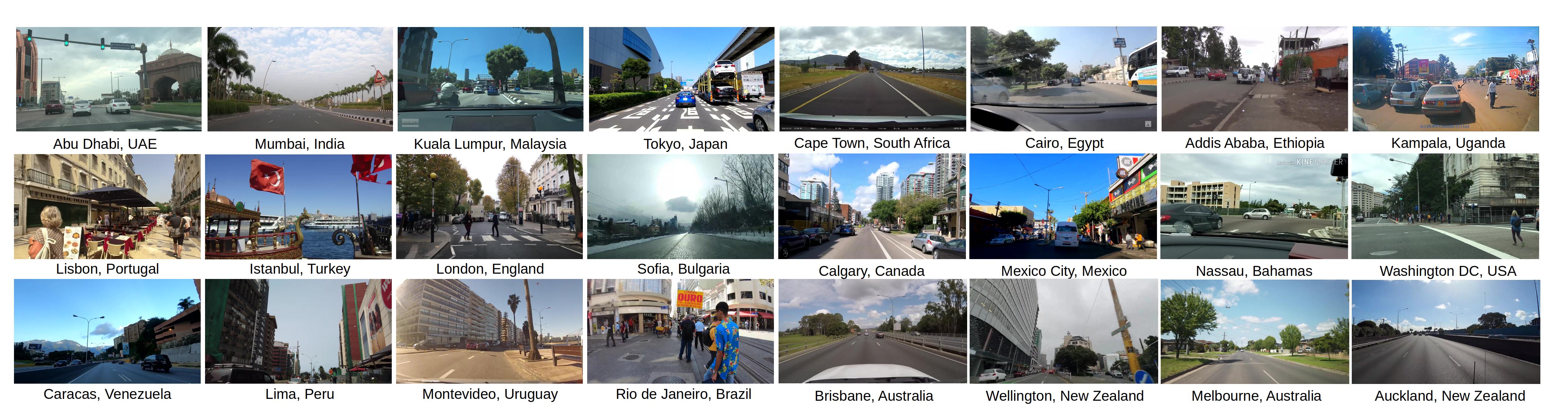}
   \caption{\textbf{Sample frames of videos from 22 different countries in the CityGuessr68k dataset.} Each quartet represents a continent. The continents in order are, Asia, Africa, Europe, North America, South America and Oceania}
   \label{fig:samples}
   \vspace{-0.25in}
\end{figure}

\noindent\textbf{Overview}
The CityGuessr$68$k dataset consists of $68,269$ first-person driving and walking videos from $166$ cities around the world. Fig. \ref{fig:samples} shows sample frames from a few videos from varied locations. Each video is annotated with hierarchical location labels, in the form of its continent, country, state/province, city.\vspace{0.05in}

\noindent\textbf{Compilation Process}
The videos in CityGuessr$68$k, pertaining to different cities, are obtained from the YouTube corpus in the form of long videos of $10$-$20$ minutes. Each video is split into numerous clips with $800$-$900$ frames, out of which $100$ frames are sampled. Each sample is then manually annotated with hierarchical labels, based on the city in which the video was recorded. City labels form the most fine-grained hierarchy, from which we go higher to state/province in which the city is located, to the country in which the state/province is located, and finally to the continent in which the country is located. Subsequently, each clip is also further divided into frames, for convenience.\vspace{0.05in}

\noindent\textbf{Post-processing}
As our dataset was collected from the YouTube corpus, and consists of a variety of driving and walking videos from numerous cities, there is a possibility that some videos might contain faces of individuals recorded at the time. To preserve the anonymity of such individuals an extensive post-processing effort was made. All the frames of all videos in the dataset were scanned with RetinaFace\cite{deng2020retinaface}. Then, scanned samples were manually inspected, after which, the detected faces were blurred maintaining the privacy of individuals. The entire procedure is described in detail in the supplementary material.\vspace{0.05in}

\begin{figure}[h]
\vspace{-0.2in}
  \centering
  \begin{subfigure}{0.48\linewidth}
    \centering
    \includegraphics[width = \textwidth]{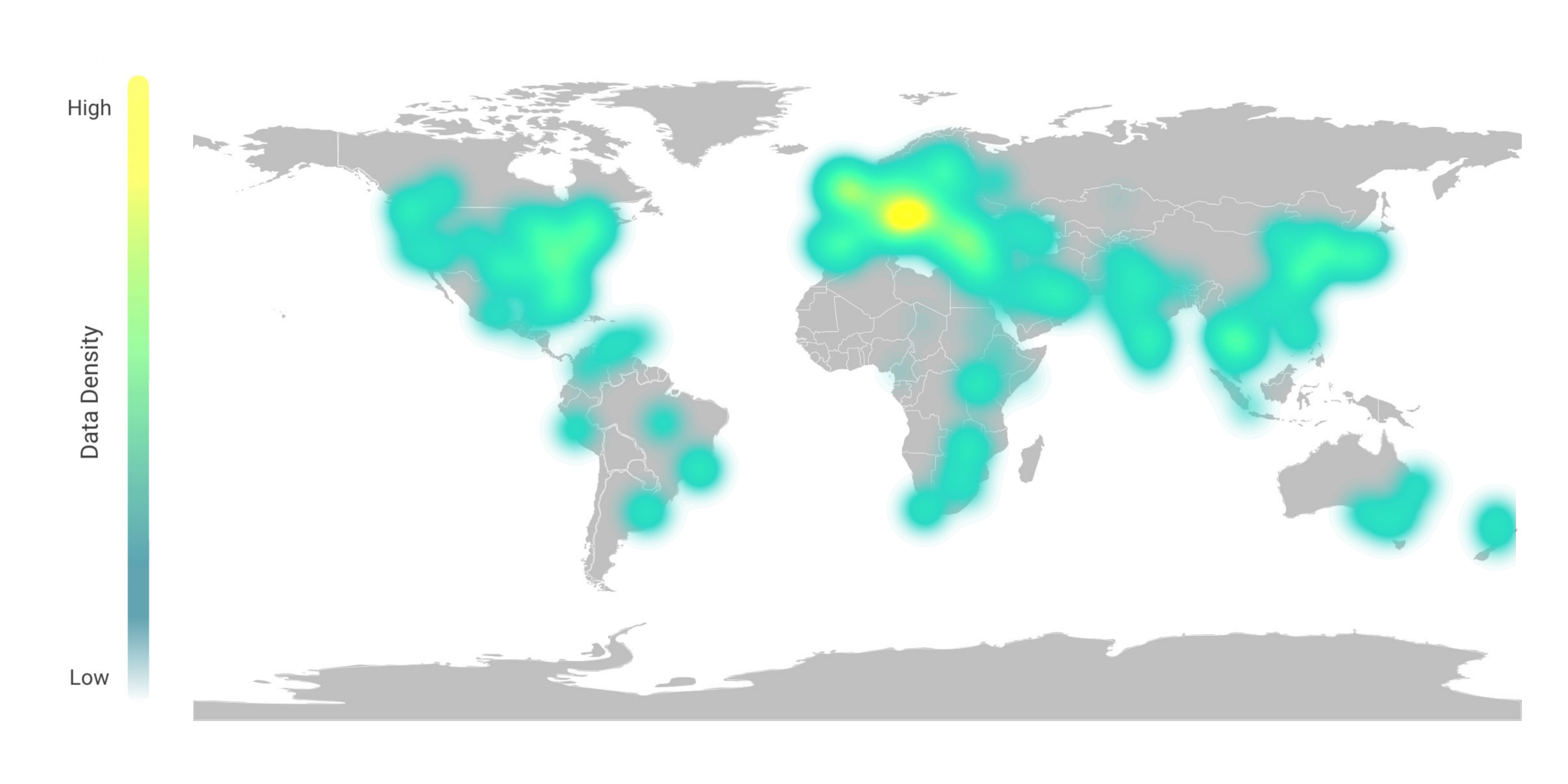}
    \caption{Distribution of videos in CityGuessr68k}
    \label{fig:ours}
  \end{subfigure}
  \begin{subfigure}{0.48\linewidth}
    \centering
    \includegraphics[width = \textwidth]{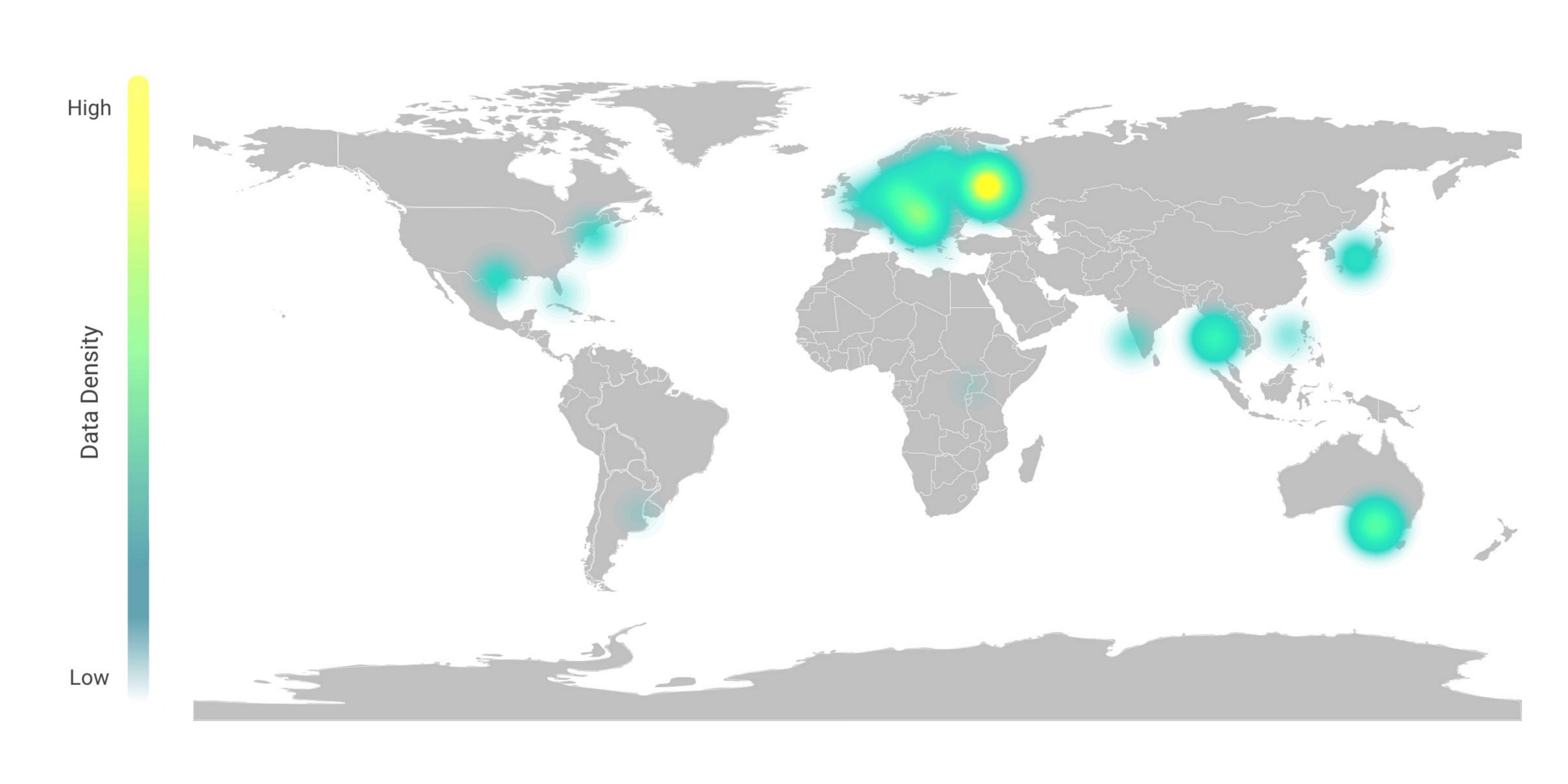}
    \caption{Distribution of videos in Mapillary(MSLS)}
    \label{fig:map}
  \end{subfigure}
   \caption{\textbf{Data distribution.} A comparison of CityGuessr$68$k with Mapillary(MSLS) dataset. CityGuessr$68$k covers more regions of the world, and has a uniform spread around the globe.}
   \label{fig:dist}
\end{figure}

\noindent\textbf{Geographical Distribution}
Our dataset consists of $68,269$ first-person driving/walking videos from $166$ cities, $157$ states/provinces, $91$ countries and $6$ continents. Fig. \ref{fig:dist} shows the distribution of data samples around the world.

\vspace{0.1in}
\begin{figure}[t]
\vspace{-0.1in}
  \centering
  \includegraphics[width = \textwidth]{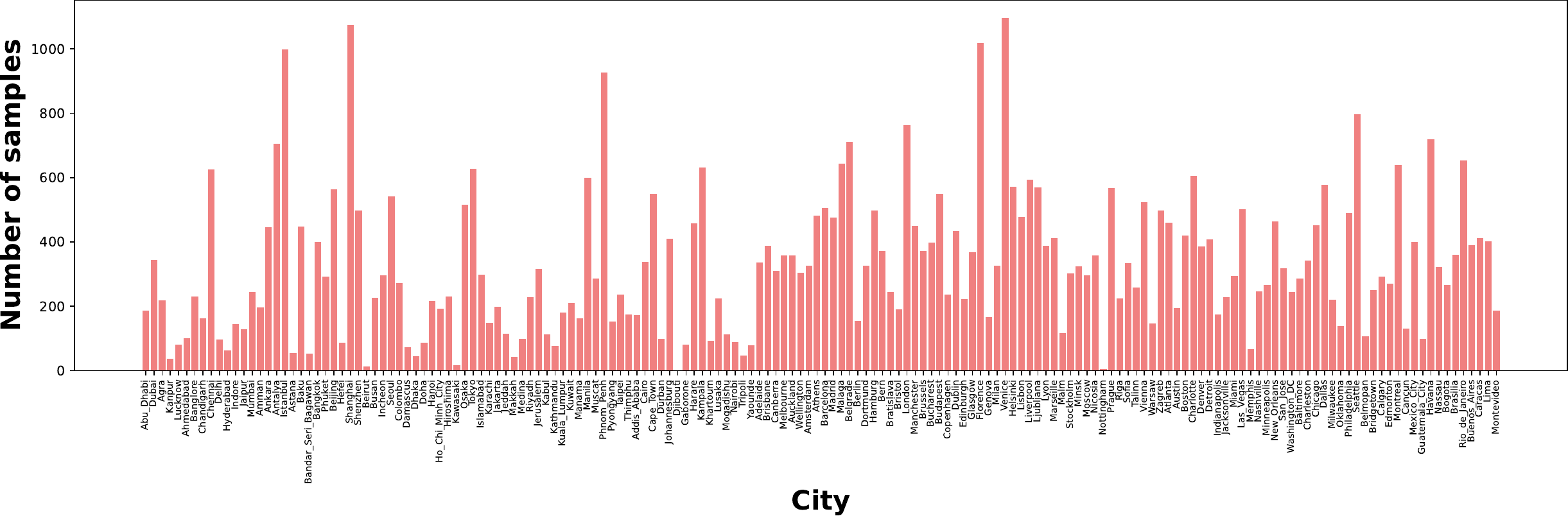}
   \caption{\textbf{Class Distribution.} Bar chart for number of samples per city class. (please zoom in for clearer class labels)}
   \label{fig:city_dist}
\end{figure}

\begin{figure}[!t]
  \centering
  \begin{subfigure}{0.24\linewidth}
    \centering
    \includegraphics[width = \textwidth]{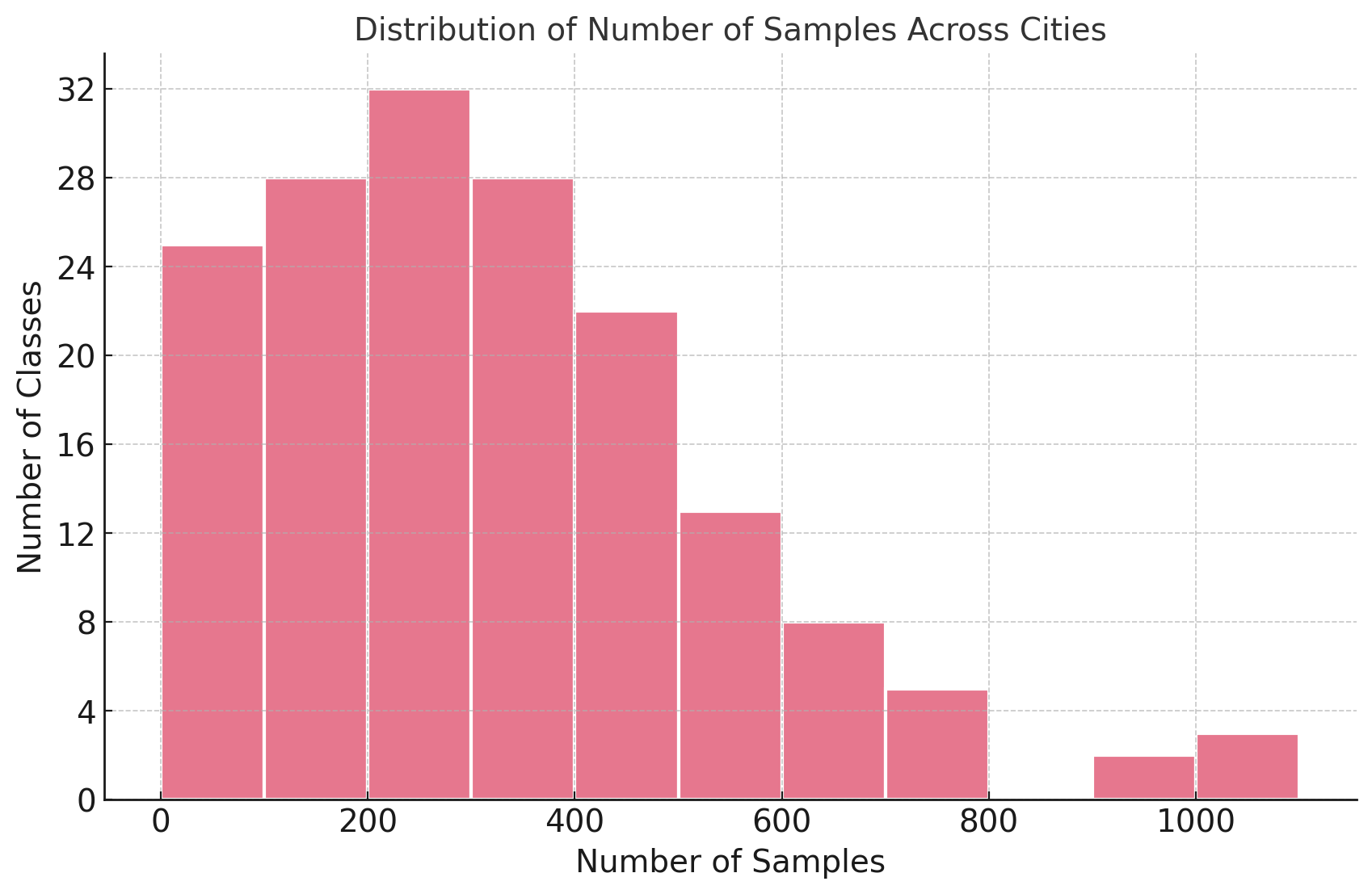}
    \caption{Histogram of city classes}
    \label{fig:city}
  \end{subfigure}
  \hfill
  \begin{subfigure}{0.24\linewidth}
    \centering
    \includegraphics[width = \textwidth]{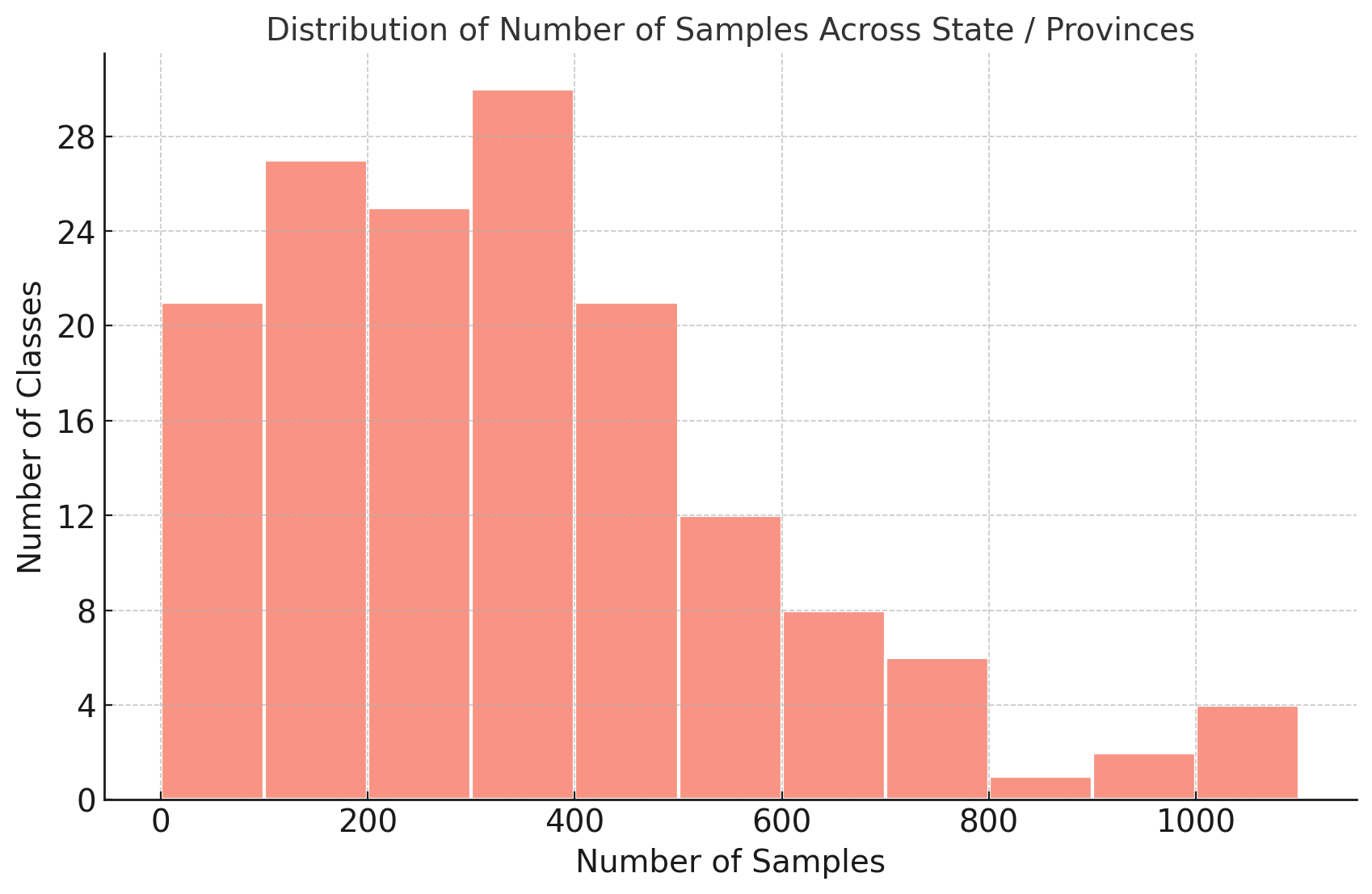}
    \caption{Histogram of state classes}
    \label{fig:state}
  \end{subfigure}
  \begin{subfigure}{0.24\linewidth}
    \centering
    \includegraphics[width = \textwidth]{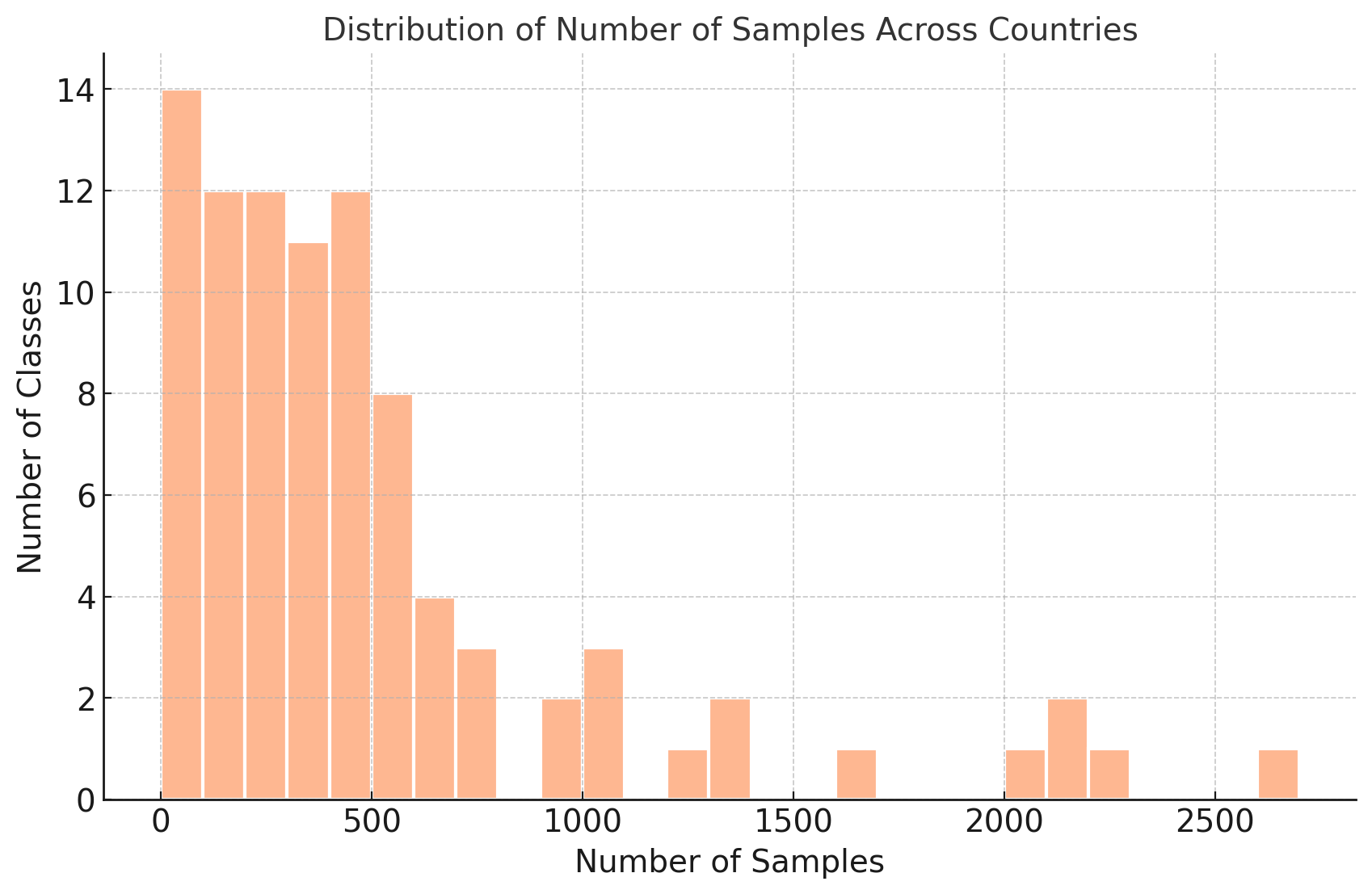}
    \caption{Histogram of country classes}
    \label{fig:country}
  \end{subfigure}
  \hfill
  \begin{subfigure}{0.24\linewidth}
    \centering
    \includegraphics[width = \textwidth]{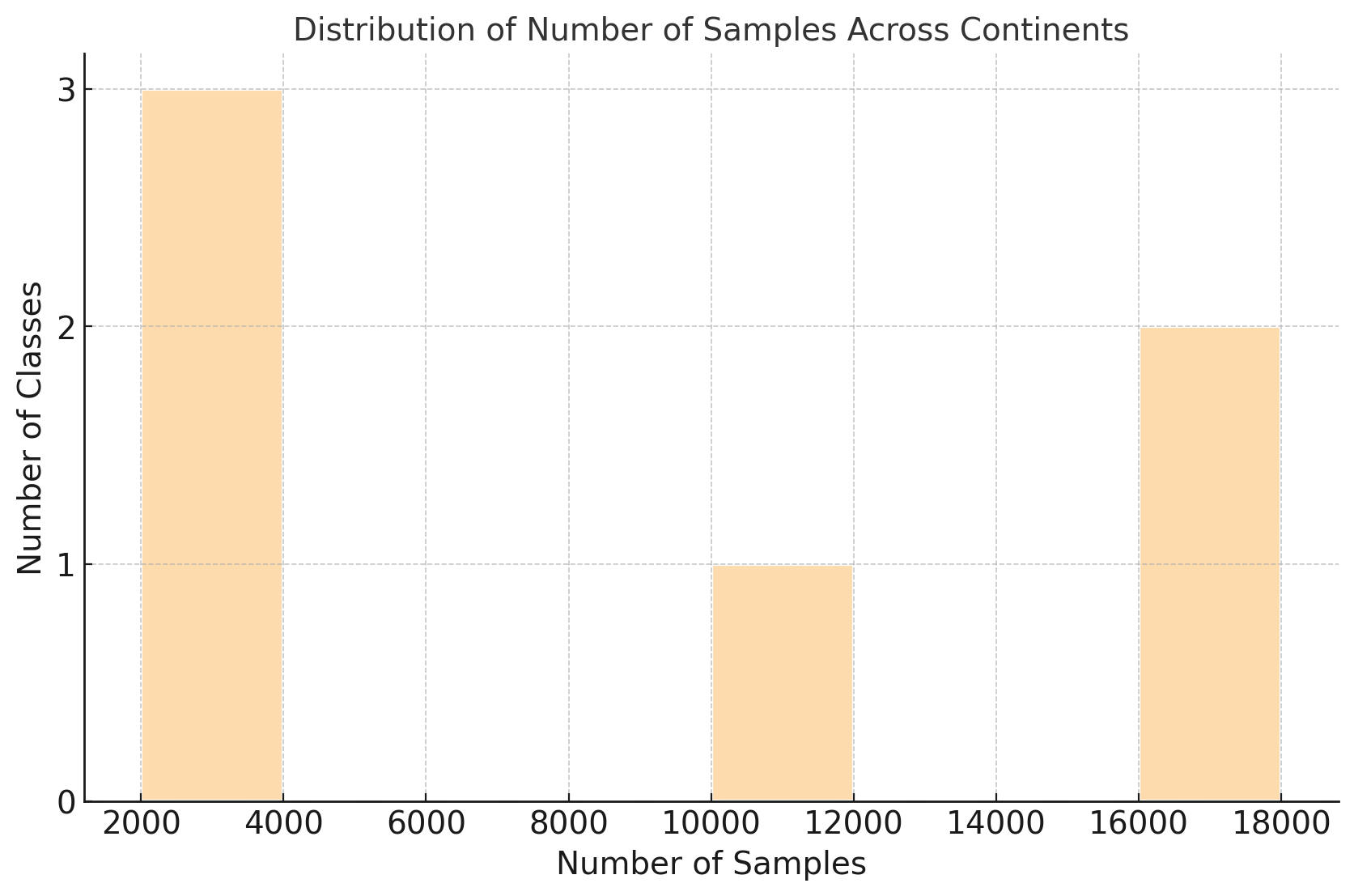}
    \caption{Histogram of continent classes}
    \label{fig:continent}
  \end{subfigure}
   \caption{\textbf{Frequency distribution.} Histograms for each hierarchy for a further statistical insight into data distribution of CityGuessr68k.}
   \label{fig:hist}
   \vspace{-0.15in}
\end{figure}

\noindent\textbf{Class Distribution}
\label{sec:cg68-class}
Each city, state/province, country and continent represents a class at their respective hierarchies. Fig. \ref{fig:city_dist} shows the class distribution of our dataset at city level. As discussed above, CityGuessr68k has a good geographical coverage. Along with that, frequency distribution among classes is also relatively even. Fig. \ref{fig:hist} shows histograms of classes at all $4$ hierarchies. Fig. \ref{fig:city} and \ref{fig:state} show that City and State classes peak in the middle of the graph with mode of both being about $250$. This means that the dataset does not have a long tail at these hierarchies. In Fig. \ref{fig:country}, we see that country classes peak very early. It may appear to have a long tail, but the height of subsequent bins is also high. As there are only $6$ continents, 3 have less than $10$k samples while $1$ has around $12$k and $2$ have around $20$k, as observed in Fig. \ref{fig:continent}, maintaining a healthy balance.\\

\noindent\textbf{Video statistics and comparison with Mapillary(MSLS)}
 Each video is divided into frames of resolution $1280$x$720$, which is higher than Mapillary(MSLS) images. All videos are approximately same in length. 
 The data is organized with every video being contained within an individual folder. Table \ref{tab:comp} shows the comparison of our CityGuessr$68$k dataset with the only other worldwide image sequence dataset, Mapillary(MSLS)\cite{warburg2020mapillary}. We see that our dataset is $\sim5$x larger and spread across more cities around the world. \\
\begin{wraptable}{r}{0.5\textwidth}
\centering
\vspace{-0.4in}
\resizebox{0.5\textwidth}{!}{
\begin{tabular}{lcc}
\hline
                           & Mapillary & CityGuessr68k \\ \hline
Number of samples          & 14,965*   & 68,269        \\
Number of cities           & 30        & 166           \\
Number of states/provinces & 29        & 157           \\
Number of countries        & 24        & 91            \\
Number of continents       & 6         & 6             \\
Consistent sequence length & No        & Yes           \\
Video sequence             & No        & Yes           \\
Uniformly Organized                  & No        & Yes           \\
Frame resolution             & 640x480        & 1280x720           \\ \hline
\end{tabular}
}
\begin{minipage}{0.5\textwidth}
\tiny *Samples with $15$ frames or more    
\end{minipage}
\caption{\textbf{Comparison with Mapillary(MSLS).} This table shows the comparison of our dataset with Mapillary. CityGuessr$68$k overcomes many shortcomings of the Mapillary dataset.}
\label{tab:comp}
\vspace{-0.4in}
\end{wraptable}

\vspace{-0.4in}
\section{Method}
\label{sec:method}
\vspace{-0.1in}
\subsection{Problem Statement and Method Overview}
\label{sec:ps}
\begin{figure}[t]
  \centering
  \includegraphics[width = \textwidth]{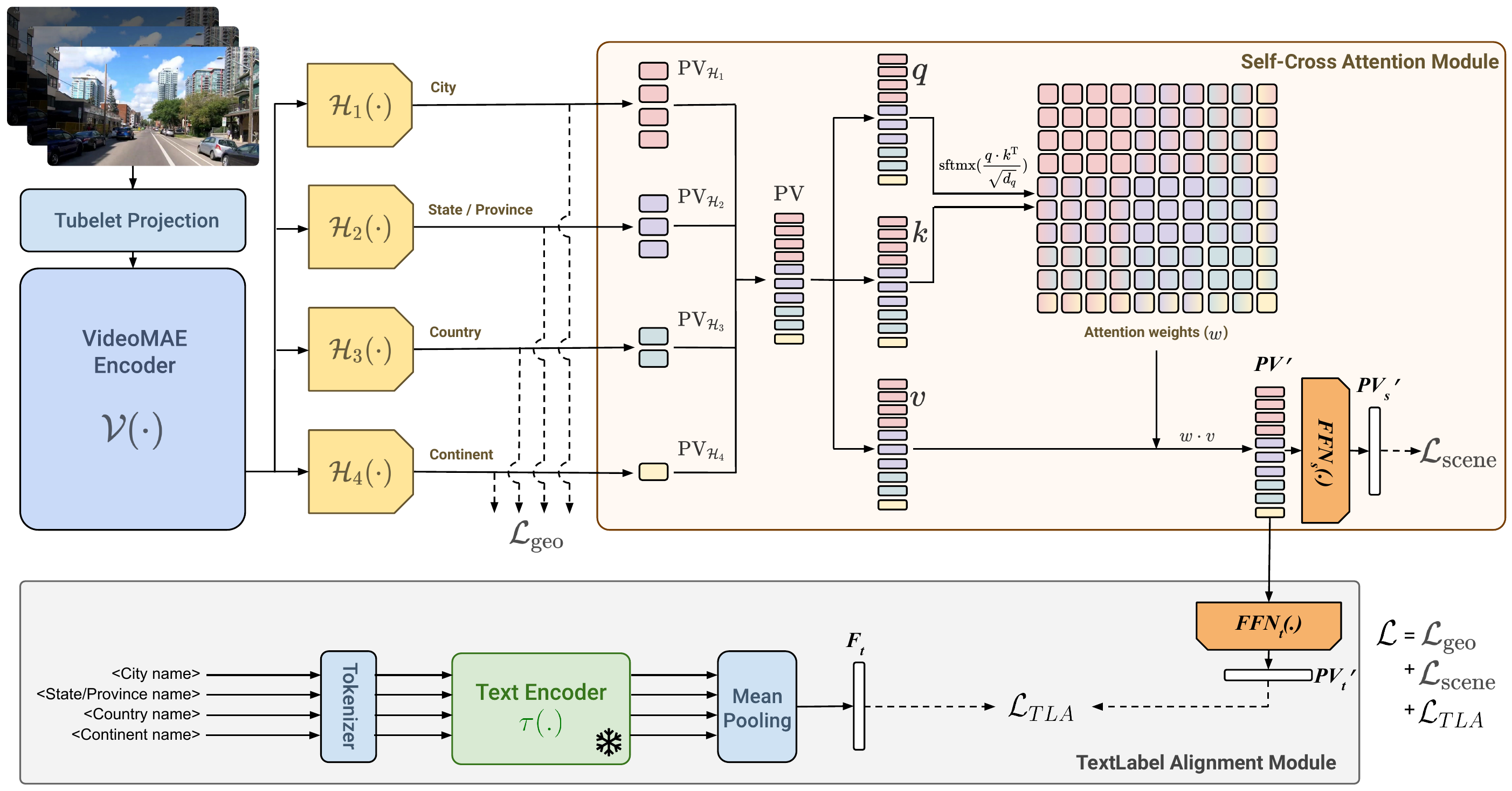}
   \caption{\textbf{Schematic Illustration of the proposed Model Architecture.} VideoMAE encoder outputs feature embeddings of the input video. The embeddings are then passed into 4 classifiers pertaining to 4 hierarchies. Their predictions are used for computing Geolocalization loss. Simultaneously prediction vectors are input into the Self-Cross Attention module, where vectors of all 4 hierarchies are concatenated and are attended to, by themselves and by each other to generate an intermediate attended vector($PV'$). In the attention weights($w$), the single colored weights along the diagonal refer to self attention weights, while the gradient double colored weights are the cross attention weights between vectors of those two different hierarchies. $PV'$ is passed simultaneously through $FFN_s$ to generate vector $PV'_s$ for Scene loss computation, and to the TextLabel Alignment module. There, it is passed through $FFN_t$ to generate vector $PV'_t$. $PV'_t$ is used for TextLabel Alignment with feature embeddings $F_t$ generated by the pretrained text-encoder from the label names of all 4 hierarchies.}
   \label{fig:method}
\vspace{-0.25in}
\end{figure}

\vspace{-0.1in}
Given an input video, our objective is to determine which city in the world, this video was recorded in. This task can also be referred to as Visual Place Recognition. Consequently, we also predict the respective state/province, country, and continent.  We approach this problem as a multi-objective classification task. Each video has a corresponding city label, state/province label, country label, and continent label. A model that solves this problem should be able to predict all these labels. Every video also has a unique scene label associated with it. We consider scene recognition and TextLabel Alignment as auxiliary tasks which aid in training our model, details of which is described in Section \ref{sec:scene} and Section \ref{sec:tla} respectively. 
\\
\textbf{Method Overview.} As shown in Fig. \ref{fig:method}, an input video is divided into tubelet tokens, as was first conceptualized in Arnab et. al\cite{arnab2021vivit}, and passed to a video encoder, which outputs feature embeddings. These embeddings are then input into $4$ classifiers, $\mathcal{H}_{1}$, $\mathcal{H}_{2}$, $\mathcal{H}_{3}$ and $\mathcal{H}_{4}$, representing the city, state/province, country, and continent predictors respectively. The outputs of these classifiers are used to compute the respective losses for each hierarchy. The classifier outputs are also passed on to our `Self-Cross Attention' module. Its outputs are simultaneously used to compute the scene loss and also passed on to the `TextLabel Alignment' Module, which aligns these features with textlabel embeddings from a pretrained text encoder via the TLA loss . All the losses (described in Section \ref{sec:loss}) are combined and backpropogated to train the model.
\vspace{-0.15in}
\subsection{Encoder Backbone}
\label{sec:backbone}
\vspace{-0.05in}
Encoder backbone is the very first stage of our model. We use VideoMAE\cite{tong2022videomae} to encode features of our input video. VideoMAE is the Masked Auto Encoder(MAE) network, first proposed in He et. al\cite{he2022masked}, adapted to videos. Originally designed for self-supervised video pretraining for recognition, VideoMAE masks a very high number of tubelets (spatiotemporal tokens), which brings performance improvement while reducing computation significantly. Thus VideoMAE becomes a very good choice for the encoder backbone. Originally VideoMAE authors use a vanilla ViT\cite{dosovitskiy2020image} backbone, and adopt the joint space-time attention\cite{arnab2021vivit}\cite{liu2021swin} to better capture high-level spatiotemporal information in the remaining tokens after masking. The VideoMAE encoder which we have incorporated, is pretrained on Kinetics-$400$\cite{kay2017kinetics} and is finetuned on our dataset. Encoded features from the backbone are then input to $4$ classifiers, one for every hierarchy, which output their respective vectors. These vectors are then passed on to the Self-Cross Attention module, which is described in the following section.
\vspace{-0.15in}
\subsection{Scene Recognition}
\label{sec:scene}
\vspace{-0.1in}
Information pertaining to each hierarchy can influence the scene of a video and vice-versa. Keeping that in mind, we aim to fuse the knowledge from all $4$ hierarchies for the identification of the scene pertaining to a video in such a way that the information relevant to each hierarchy enhances the knowledge of other hierarchies as well as its own. This is only possible if there is a means for hierarchies to interact with one another. To this end, we propose 
Self-Cross Attention module, as described below.
\vspace{-0.2in}
\subsubsection{Self-Cross Attention module}
\label{sec:selfcross}
As mentioned above, scene predictions can be influenced by all hierachies and they need to interact with each other to enhance scene recognition. Our Self-Cross Attention module is designed for that purpose. As shown in Fig. \ref{fig:method}, the Self-Cross Attention module takes the output vectors of all $4$ hierarchies ($PV_{\mathcal{H}_1} \in \mathbb{R}^{d_1}$, $PV_{\mathcal{H}_2} \in \mathbb{R}^{d_2}$, $PV_{\mathcal{H}_3} \in \mathbb{R}^{d_3}$, $PV_{\mathcal{H}_4} \in \mathbb{R}^{d_4}$, where $d_1, d_2, d_3, d_4$ are the number of classes in city, state/province, country and continent hierarchies respectively). These vectors are concatenated to form a vector $PV  = concat(PV_{\mathcal{H}_1}, PV_{\mathcal{H}_2},PV_{\mathcal{H}_3},PV_{\mathcal{H}_4})\in \mathbb{R}^{d}$($d = d_1 + d_2 + d_3 + d_4$). Then the vector is projected into query($q$), key($k$) and value($v$) vectors, which are then used to compute multihead attention\cite{vaswani2017attention} (defined as $softmax(\frac{qk^T}{\sqrt{d_q}}).v$, where $d_q$ is the query dimension), $PV' = MHA(PV)$. The attended output is again projected into a vector $\in \mathbb{R}^{d}$, which is finally passed through a feed-forward network ($FFN_s$), to gradually reduce the dimension and output the scene vector ($PV'_s = FFN_s(PV')\in \mathbb{R}^{d_s}$, where $d_s$ is the number of scene labels). $PV'_s$ is then used to compute scene loss.
Thus, the module essentially performs self attention on output of each hierarchy classifier, as well as cross attention between each pair of hierarchies as depicted in Fig. \ref{fig:method}, conceiving the name, \textit{Self-Cross Attention} module. This procedure achieves the intended effect of the outputs from all hierarchies interacting with each other, and with themselves, enhancing location prediction. $PV'$ is also passed on to TextLabel Alignment module.
\vspace{-0.2in}
\subsubsection{Scene labels}
\label{sec:soft}
The Self-Cross Attention module, discussed in the previous section, outputs a Scene Vector($PV'_s$), which gives a prediction for the scene. To compute scene loss, we require scene labels corresponding to each video sample. For images, assigning scene labels is straightforward, but the same isn't the case with videos. Assigning one scene label to an entire video involves a lot of nuance, as 
scene might change as we go through all frames. The most trivial way of scene labelling a video can just involve taking the first/middle/random frame and assigning the scene label of that frame to the entire video. That is not a good direction to pursue, as the scene of that frame might not represent the video properly. Majority voting is a more clinical way of approaching this task. This involves assigning a scene label to all frames of a video and assign the scene label which represents the most amount of frames to the video. This is indeed better than the previous approach, but it, too, does not fully capture the complete variation of the video. 

Thus, we devise a simple yet effective way of representing the scene label of a video. The concept of soft-labels is interesting, as it captures the detail of a representation. Generally, soft labels are used to train teacher-student distillation models\cite{hinton2015distilling}. Instead of assigning a definite class to a sample, soft labels use percentages which capture the probabilities of the sample belonging to each class. Soft labels fit perfectly into our problem setup, as each value of class can be represented as the percentage of frames in the video that belong to a certain class. We use this technique to provide a proper representation of the scene of a video, and assign soft-labels to all videos as the ground truth, which are then used for computing scene loss.

For assigning scene labels to video frames, we use the labels provided by Places$2$\cite{zhou2017places} dataset. We implement a pre-trained scene classification model provided in \cite{zhou2017places}, and get labels for all frames. Then, we convert those into soft labels as mentioned above. We use $16$ image scene labels for this work.
\subsection{TextLabel Alignment Strategy}
\label{sec:tla}
Associating a name with a picture/video of a location is helpful for humans to retain and identify the characteristics of that location. Thus location names can be instrumental assets for training a geolocalization model. We pursue this by distilling knowledge from these labels into our model in feature space, through our TextLabel Alignment (TLA) Strategy.
\vspace{-0.2in}
\subsubsection{Strategy}
We compute the textual features of the location labels via a pretrained text-encoder\cite{radford2021learning}. 
The objective 
is to align the features of our model 
with the generated textlabel features. As is the case with scenes, alignment of textlabel features with combined features of all hierarchies is essential for maximum benefit. Thus we 
use the features output from the Multihead Attention layer of the Self-Cross Attention module (Section \ref{sec:selfcross}), before it is passed on to $FFN_s$. The attended vector ($PV'$ in Section \ref{sec:scene}) is passed through a different feed-forward network ($FFN_t$) to obtain an output vector($PV'_t = FFN_t(PV') \in \mathbb{R}^{d_t}$, where $d_t$ is the textlabel feature dimension).
$PV'_t$ is then used to compute the TLA loss (Eq. 3).
Thus, the attended features of all hierarchies are aligned with the textlabel features generated from the text encoder. 
This helps to 
distill knowledge from the textlabels to the model features 
by associating the location's name with the respective video, leading to further enhancement of location prediction.
\vspace{-0.2in}
\subsubsection{TextLabel Feature Computation}
As mentioned above, TextLabel features can be computed in multiple ways. The most trivial way of approaching this is passing only the city names through the text encoder, as association of the city with the input video is the most essential by virtue of it being the finest hierarchy. 
However, we do lose 
information from other coarser hierarchies. 
If all hierarchies are to be incorporated, we can pass each hierarchy label through the text encoder separately and combine the output of the embeddings to obtain the final TextLabel features. The latter option allows us to consider input from all hierarchies without compromising the expression of each hierarchy. 
A detailed empirical performance analysis with each alternative is shown in Section \ref{sec:res-tla}.
\vspace{-0.1in}
\subsection{Losses}
\label{sec:loss}
\vspace{-0.05in}
As evident in Fig. \ref{fig:method}, our network is trained with three losses, geolocalization loss($L_{geo}$), scene loss($L_{scene}$) and TLA loss($L_{TLA}$). Geolocalization loss is computed as a combination of Cross-Entropy losses(CE) of each hierarchy. Given a video V, $L_{geo}$ can be defined as,
\vspace{-0.1in}
\begin{align}
    L_{geo}(V) = \sum_{i \in \{city,state,country,continent\}} CE(l_i, \hat{l}_i),
\vspace{-0.3in}
\end{align}
where, $l$ denotes the ground truth label, while $\hat{l}$ denotes the predicted label. Scene loss is computed as the cross entropy loss between the soft label assigned to a video, and the output vector from the Self-Cross Attention module.
\vspace{-0.1in}
\begin{align}
    L_{scene}(V) = CE(s, PV'_s),
\vspace{-0.15in}
\end{align}
where, $s$ denotes the ground truth soft-scene label, and $PV'_s$ denotes the predicted scene vector. TLA loss is computed as the negative cosine similarity between the text features generated from the class labels, and the output vector to be aligned.
\vspace{-0.15in}
\begin{align}
    L_{TLA}(V) = -Cosine\_Similarity(F_t, PV'_t),
\vspace{-0.15in}
\end{align}
where, $F_t$ denotes the text features from class labels and $PV'_t$ denotes the output vector. Thus the total loss is
\vspace{-0.1in}
\begin{align}
    L(V) = L_{geo}(V) + L_{scene}(V) + L_{TLA}(V).
\vspace{-0.15in}
\end{align}
\vspace{-0.3in}
\subsection{Inference}
\label{sec:inf}
\vspace{-0.05in}
We geolocalize the video V with the outputs of the four classifiers $\mathcal{H}_1$, $\mathcal{H}_2$, $\mathcal{H}_3$ and $\mathcal{H}_4$. $\mathcal{H}_4$ predicts the continent label, $\mathcal{H}_3$ predicts the country label, $\mathcal{H}_2$ predicts the state/province label and $\mathcal{H}_1$ predicts the city label. Now, the predictions for fine-grained hierarchies could be improved with the assistance of the coarser hierarchies. To refine the probabilities of a hierarchy prediction, we can multiply the probabilities of the coarser hierarchies, as they would push the probabilities of the classes in which the coarser hierarchies are most confident. Thus
\vspace{-0.1in}
\begin{align}
    P({C_i}^{\mathcal{H}_1}|V) = P({{C_i}^{\mathcal{H}_1}|V}) * P({{C_j}^{\mathcal{H}_2}|V}) * P({{C_k}^{\mathcal{H}_3}|V}) * P({{C_l}^{\mathcal{H}_4}|V})
\vspace{-0.2in}
\end{align}

\noindent where city ${C_i}^{\mathcal{H}_1}$ is located in state ${C_j}^{\mathcal{H}_2}$, which is located in country ${C_k}^{\mathcal{H}_3}$, which in turn is located in continent ${C_l}^{\mathcal{H}_4}$. Continent probabilities are multiplied into countries and so on. After performing these operations, the predictions of individual probabilities could be independent of each other, or could be codependent on each other. Codependent predictions could be performed by first predicting the most fine-grained hierarchy after multiplying probabilities, and then tracing the hierarchical structure upwards, i.e., once the city is predicted, the state prediction would simply be the state in which the city is located, and so on. Independent predictions are just made by using the individual hierarchy probabilities to determine the most likely class. In our final method, we use codependent predictions, as we found that empirically they are more accurate. More analysis on that is covered in Section \ref{sec:abl-eval}.
\vspace{-0.1in}
\section{Experiments, Results and Discussion}
\label{sec:exp}
\vspace{-0.1in}
This section describes the details of the experiments performed with our model, the data used for those experiments and the training and validation setup in regards to the same.
\vspace{-0.15in}
\subsection{Data and metrics}
\label{sec:data}
\vspace{-0.05in}
For training and validating our model, we use our newly proposed dataset CityGuessr$68$k. The dataset is divided into an $80$:$20$ stratified train-test split, all classes being represented in both sets. Thus we train our model on $54,614$ videos and validate on $13,655$ videos. We assess our model's performance using prediction accuracy (top1 accuracy), for each hierarchy. We show additional analysis on top5 accuracy in the supplementary material.
\vspace{-0.2in}
\subsection{Training Details}
\label{sec:train}
\vspace{-0.05in}
Our model is implemented in pytorch\cite{paszke2019pytorch}. The input video consists of $15$ frames, resized to $224$x$224$. The model is trained on one node of an NVIDIA RTX A$6000$ GPU. The VideoMAE version used has a ViT-S\cite{dosovitskiy2020image} backbone, and its weights are pretrained on Kinetics-$400$\cite{kay2017kinetics}. The model is trained for $10$ epochs with a batch size of $12$. We use the Adam\cite{kingma2014adam} optimizer with a learning rate of $0.001$. The Self-Cross Attention module has $2$ heads, an embedding dimension of $6$ and an FFN that has $6$ layers. The TextLabel Alignment strategy utilizes an FFN with $3$ layers and the text embedding feature dimension is $512$. The video encoding feature dimension is $384$. 
\vspace{-0.2in}
\subsection{Utility of Video data}
\label{sec:res-imgvsvid}
\vspace{-0.1in}
Video-based geolocalization is more accurate than using single images because videos contain more richer information. We carry out experiments where we replace our video backbone in with its image counterpart (MAE\cite{he2022masked}), keeping all other details exactly the same. We test four different settings for the image model. As shown in Table \ref{tab:imgvsvid} we observe that the ``random" setting performs the best. We also observe that the video model outperforms the image models by a large margin. it achieves a $9\%$ improvement over the best image model setting, showing the utility of video data. \\

\noindent
\begin{minipage}[c]{0.48\textwidth}
\centering
\resizebox{\textwidth}{!}{
\begin{tabular}{clcccc}
\hline
\multicolumn{1}{l}{\textbf{Backbone}} & \textbf{Setting}      & \textbf{City} & \textbf{State} & \textbf{Country} & \textbf{Continent} \\ \hline
\multirow{4}{*}{MAE}         & First frame  & 52.1 & 52.6           & 55.3    & 70.4      \\
                             & Mid frame    & 48.9 & 49.3           & 54.6    & 69.8      \\
                             & Last frame   & 48.1 & 48.4           & 53.4    & 69.3      \\
                             & Random frame & 55.8 & 56.3           & 60.8    & 74.1      \\ \hline
\multicolumn{1}{l}{VideoMAE} & video        & 64.5 & 64.5           & 65.9    & 74.4      \\ \hline
\end{tabular}
}
\captionof{table}{\textbf{Comparison of image and video backbones.} Comparing performance of MAE and VideoMAE models to demonstrate the necessity of video geolocalization.}
\label{tab:imgvsvid}
\end{minipage}
\hfill
\begin{minipage}[c]{0.48\textwidth}
\centering
\resizebox{\textwidth}{!}{
\begin{tabular}{lccccc}
\hline
\multicolumn{1}{c}{\textbf{Scenes}} & \textbf{TLA}                                                         & \textbf{City} & \textbf{State} & \textbf{Country} & \textbf{Continent} \\ \hline
\multicolumn{1}{c}{-}               & -                                                                    & 64.5          & 64.5                    & 65.9             & 74.4               \\
Majority                            & -                                                                    & 66.9          & 67.3                    & 72.1             & 81.1               \\
\rowcolor{lightgray} 
Soft                                & -                                                                    & 67.9          & 68.4                    & 72.4             & 81.6               \\ \hline
Soft                                & \multicolumn{1}{l}{city only}                                        & 69.1          & 69.5                    & 73.7             & 83.1               \\
\rowcolor{lightgray}
Soft                                & all hierarchies & \textbf{69.6} & \textbf{70.2}           & \textbf{74.8}    & \textbf{83.8}      \\ \hline
\end{tabular}
}
\captionof{table}{\textbf{Effect of adding Scene recognition and TextLabel Alignment(TLA).} Comparing performance of the model with variants of scene labels and TLA}
\label{tab:scene}
\end{minipage}

\subsection{Scene Recognition}
\label{sec:res-scene}
\vspace{-0.1in}
We have introduced a novel technique for incorporating scenes to aid our model training. Self-Cross attention(Section \ref{sec:selfcross}) between the prediction vectors of hierarchies was conceptualized in an attempt to improve location predictions. We also devised an elegant way of representing scenes with soft labels(Section \ref{sec:soft}) which gives a proper representation of a video scene label making it more suitable for loss computation. Table \ref{tab:scene} shows that addition of Self-Cross Attention module certainly helps the model to train better and gives better validation performance. We also showcase our results on two variations of scene labels, one obtained by majority voting and other with soft labels. Comparing their performance, we see that soft labels are 
more helpful in model training. Note that both models use hierarchical evaluation with codependent predictions.
\vspace{-0.1in}
\subsection{Benefit of TextLabel Alignment}
\label{sec:res-tla}
\vspace{-0.1in}
We employ our TextLabel Alignment strategy(Section \ref{sec:tla}) to distill knowledge from the names of the locations into our model in the feature space, by aligning textlabel features generated by the pretrained text encoder to the features of our model. We described 2 strategies for computing textlabel features in Section \ref{sec:tla}, from city labels, and from mean of features from all hierarchy labels. Table \ref{tab:scene} shows that incorporation of the TextLabel Alignment strategy enhances the features of the model, thus giving a better performance. We showcase our results on both the above described variations. Comparing their performance, we see that using all hierarchies helps the model train better as hypothesized. Note that both models use hierarchical evaluation with codependent predictions.
\subsection{Independent v/s Codependent Hierarchical Evaluation}
\label{sec:abl-eval}
As discussed in Section \ref{sec:inf}, hierarchical evaluation enhances the predictions of the model and as a consequence, improving geolocalization performance. Table \ref{tab:he} shows the same. After multiplying probabilities, coarser hierarchy predictions could be independent, or they could be codependent on finer hierarchies. We evaluate the model using both variations, and 
Table \ref{tab:he} shows that codependent predictions are better than independent predictions. Note that all the results include Self-Cross Attention module with soft-scene labels and TextLabel Alignment with all hierarchies.\\

\noindent
\begin{minipage}[c]{0.48\textwidth}
\centering
\resizebox{\textwidth}{!}{
\begin{tabular}{lcccc}
\hline
\textbf{Model}         & \multicolumn{1}{l}{\textbf{City}} & \multicolumn{1}{l}{\textbf{State}} & \multicolumn{1}{l}{\textbf{Country}} & \multicolumn{1}{l}{\textbf{Continent}} \\ \hline
w/o hierarchical eval. &        69.1                      &             69.6                          &     72.5                             &        79.2                          \\
Independent            &      69.6                       &         69.8                               &       72.5                          &              79.2                    \\
\textbf{Codependent}   & \textbf{69.6} & \textbf{70.2}           & \textbf{74.8}    & \textbf{83.8}                         \\ \hline
\end{tabular}
}
\captionof{table}{\textbf{Hierarchical Evaluation.} Comparing variants of the model with different types of hierarchical evaluation techniques} 
\label{tab:he}
\end{minipage}
\hfill
\begin{minipage}[r]{0.48\textwidth}
\centering
\resizebox{\textwidth}{!}{
\begin{tabular}{lcccc}
\hline
\textbf{Model}                  & \textbf{City} & \textbf{State} & \textbf{Country} & \textbf{Continent} \\ \hline
PlaNet\cite{weyand2016planet}                          & 55.8          & 56.3                    & 60.8             & 74.1               \\
ISNs\cite{muller2018geolocation}                   &   59.5      &      59.9         &          64.1               &     75.9                                 \\
GeoDecoder\cite{clark2023we}                      & 64.2          & 64.5                    & 69.5             & 79.9               \\ \hline
Timesformer\cite{bertasius2021space} & 60.9          & 61.4                    & 66.1             & 78.4               \\
VideoMAE\cite{tong2022videomae}                        & 64.5          & 64.5                    & 65.9             & 74.4               \\ \hline
\textbf{Ours}                   & \textbf{69.6} & \textbf{70.2}           & \textbf{74.8}    & \textbf{83.8}      \\ \hline
\end{tabular}
}
\captionof{table}{Comparison of our method with baselines and state-of-the-art methods}
\label{tab:res}
\end{minipage}
\vspace{-0.1in}
\subsection{Comparison with State-of-the-art}
\label{sec:sota}
\vspace{-0.1in}
As stated in Section \ref{sec:data}, we show the validation performance of our model on $13,655$ videos from CityGuessr$68$k. As no worldwide video geolocalization methods exist, we compare our model to the baselines with TimesFormer\cite{bertasius2021space} and VideoMAE\cite{tong2022videomae} encoders, along with the relevant state-of-the-art image geolocalization methods. For image models, we use the random frame setting, as it performed the best for the image MAE\cite{he2022masked} baseline(Section \ref{sec:res-imgvsvid}). Hierarchy classifiers are included for all models, and everything else is kept the same as per specifications mentioned in Section \ref{sec:train}. Table \ref{tab:res} shows the results of our model on our dataset. Our model is able to achieve a $69.6\%$ top1 accuracy on City prediction, i.e., the most fine-grained hierarchy. Our model showcases an improvement of $\sim6\%$ highlighting the significance of our modules. Our model also shows an improvement in the coarser hierarchies with an $\sim6\%$ jump in state/province prediction, an $\sim5\%$ improvement in country and a $\sim4\%$ in continent prediction. 
\vspace{-0.3in}
\section{Performance on Mapillary(MSLS)}
\label{sec:map}
\vspace{-0.1in}
Mapillary(MSLS)~\cite{warburg2020mapillary} is an image sequence dataset, with sequences of varying length from $30$ cities around the world. As discussed in Section \ref{sec:cg68}, Mapillary does have some shortcomings. Also, from Fig. \ref{fig:map}, we observe that Mapillary does not cover a lot of locations around the world. We propose CityGuessr$68$k to address all these concerns. 
However, we 
validate the effectiveness of our model by also showing its performance on Mapillary.

\noindent\textbf{Data preparation}
Due to its design, there are a lot of steps involved in making the Mapillary dataset compatible with our model. Mapillary is an image sequence dataset spread across multiple folders and subfolders. To its credit, it was originally collected for training image sequence retrieval models and thus every city has some query sequences and some database sequences. As we are performing a classification task, we do not require seperate query and database sequences. Thus we decided to combine sequences from both for our purposes. We also further reformat, filter and split the dataset such that it is compatible with our problem configuration. The procedure of data preparation is further detailed in the Supplementary material.
\begin{wraptable}{r}{0.5\textwidth}
\centering
\vspace{-0.3in}
\resizebox{0.5\textwidth}{!}{
\begin{tabular}{lcccc}
\hline
\textbf{Model} & \multicolumn{1}{l}{\textbf{City}} & \multicolumn{1}{l}{\textbf{State}} & \multicolumn{1}{l}{\textbf{Country}} & \multicolumn{1}{l}{\textbf{Continent}} \\ \hline
VideoMAE        & 67.6                              & 67.6                                        & 68.2                                & 81.9                                  \\
\textbf{Ours}  & \textbf{72.8}                    & \textbf{72.8}                              & \textbf{73.2}                       & \textbf{88.1}                         \\ \hline
\end{tabular}
}
\caption{\small{\textbf{Performance comparison on Mapillary(MSLS) dataset.} Our method compared with the VideoMAE baseline}}
\label{tab:map}
\vspace{-0.25in}
\end{wraptable}
\noindent\textbf{Experiments and Results}
\label{sec:map-exp}
After the filtering and train-test split, we had $9049$ train sequences and $2271$ validation sequences. We assess the performance again using prediction accuracy (top$1$ accuracy). Training parameters were kept exactly the same as described in Section \ref{sec:train}. Table \ref{tab:map} shows the validation results of our model with Self-Cross Attention module trained with soft-scene labels, TextLabel Alignment strategy with all hierarchies and codependent hierarchical evaluation. We compare the performance of our model against the VideoMAE baseline. We see that there's a $5\%$ jump in top1 accuracy on city prediction, as well as significant improvements in coarser hierarchies as well. This shows that our model trains and performs well on other datasets and thus can be generalized for this task across different data distributions.
\vspace{-0.1in}

\section{Conclusion}
\label{sec:conc}
\vspace{-0.05in}
In this paper, we formulated a novel problem of worldwide video geolocalization. As there is no large scale dataset 
to tackle this challenging problem, we introduced a new global level
video dataset, CityGuessr68k, containing 68,269 videos from 166 cities. We also proposed a baseline approach which consists of a transformer-based architecture with a Self-Cross Attention module for incorporating an auxiliary task of scene recognition with soft-scene labels as well as a TextLabel Alignment strategy to distill knowledge from location labels in feature space. We demonstrated the efficacy of our method on our dataset as well as on Mapillary(MSLS) dataset. 
As a future direction, we plan to explore the generalizability of the combination of Self-Cross Attention module and TextLabel Alignment to other hierarchical video classification tasks.

\section*{Acknowledgements}
This work was supported in parts by the US Army contract  W911NF-2120192 and National Geospatial-Intelligence Agency(NGA) Award \# HM0476-20-1-0001. We would like to extend our gratitude to all the reviewers for their valuable suggestions. We like to thank high school students Emily Park, Megan Shah and Vikram Kumar for their contributions towards collecting data and to Robert Browning and Dr. Krishna Regmi for mentoring them. We would also like to thank Vicente Vivanco Cepeda, Akashdeep Chakraborty, Prakash Chandra Chhipa, Manu S Pillai and Brian Dina for their contributions towards the dataset and insightful discussions.
%
%
\bibliographystyle{splncs04}
\bibliography{main}

\title{CityGuessr: City-Level Video Geo-Localization on a Global Scale (Supplementary Material)} 

\titlerunning{CityGuessr}

\author{Parth Parag Kulkarni\inst{1} \and
Gaurav Kumar Nayak\inst{2} \and
Mubarak Shah\inst{1}}

\authorrunning{Kulkarni et al.}

\institute{Center for Research in Computer Vision. University of Central Florida, USA \and
Mehta Family School of DS \& AI, Indian Institute of Technology Roorkee, India
\email{parthparag.kulkarni@ucf.edu; gauravkumar.nayak@mfs.iitr.ac.in; shah@crcv.ucf.edu}}

\maketitle

\vspace{0.2in}
In this supplementary material, we provide additional insights 
on our method and dataset. 
We organize our supplementary into the following sections: 
\vspace{0.2in}
\begin{enumerate} 
\setlength\itemsep{1.5em}
    \item Ablation on Number of Scenes
    \item Ablation on Text Encoders
    \item Top5 Analysis
    \item Data preparation for Mapillary(MSLS)
    \item Post processing procedures for Anonymity preservation
    \item Analysis of the CityGuessr68k Dataset
   \vspace{0.15in}
    \begin{enumerate}
    \setlength\itemsep{1.5em}
        \item Additional Visualizations of Data Samples
        \item Lorenz Curves
        \item Gini Coefficient
        \item Hoover Index
    \end{enumerate}
    \item Qualitative Results 
    \vspace{0.15in}
    \setlength\itemsep{1.5em}
    \begin{enumerate}
        \item Correct Predictions Around the World
        \item Examples of Localizations
    \end{enumerate}
\end{enumerate}
\newpage
\section{Ablation on Number of Scenes}
\label{sec:abl-noscene}
Our method constitutes of an auxiliary task of scene recognition for better model training which enables 
more accurate location prediction of a video. As described in Section \textcolor{red}{4.3} in the main paper, we introduce a self-cross attention module to extract and combine information from all hierarchies in a meaningful manner. We also incorporate soft-scene labels to further improve the performance. Table \textcolor{red}{3} of the main paper provides quantitative evidence for the same.

We used the labels provided by Places2~\cite{zhou2017places} dataset for assigning scene labels to video frames (main paper Section \textcolor{red}{4.3}), which are converted to soft labels. As mentioned, we used $16$ scene labels. The labeling structure of Places2 is hierarchical in nature, and under these 16 labels, there are 365 finer labels, pertaining to different subcategories of scenes within each category.  For eg., the finer scene labels like `/a/airplane\_terminal', `/c/car\_interior', `/c/cockpit', `/t/train\_interior' come under the umbrella of `transportation (vehicle interiors, stations, etc.)' which is one of the $16$ coarser labels. 

Table \ref{tab:number-scene} shown below provides an ablation study in which we compare the performance 
of our model with $16$ and $365$ scene labels. As the focus of this analysis was exclusively on scenes, we have trained the models using only Eq. \textcolor{red}{1} and \textcolor{red}{2} of the main paper. We observe that 365 labels could not improve the performance further in comparison to the model trained with 16 labels. This might imply that the model does not benefit from the extremely fine-grained nature of 365 scene labels, which justifies the use of 16 scene labels in our model. 

\begin{table}[!h]
\centering
\vspace{-0.2in}
\resizebox{0.7\textwidth}{!}{
\begin{tabular}{ccccc}
\hline
\textbf{Number of Scenes}         & \multicolumn{1}{l}{\textbf{City}} & \multicolumn{1}{l}{\textbf{State}} & \multicolumn{1}{l}{\textbf{Country}} & \multicolumn{1}{l}{\textbf{Continent}} \\ \hline
\textbf{16}      & \textbf{67.9}                              & \textbf{68.4}                                        & \textbf{72.4}                                & \textbf{81.6}                                  \\
365      & 67.6                              & 68.1                                        & 72.1                                & 81.2                                  \\
\hline
\end{tabular}
}
\vspace{0.05in}
\caption{\textbf{Ablation Study on Number of Scenes.} Comparing variants of the model trained with different number of scene labels. We show that using 365 scenes does not improve the performance and 16 scene labels work the best}
\vspace{-0.5in}
\label{tab:number-scene}
\end{table}
\section{Ablation on Text Encoders}
\label{sec:abl-tenc}
In Section \textcolor{red}{4.4} in the main paper, we discussed our TextLabel Alignment(TLA) strategy, which aligns model features with textlabels from all hierarchies, i.e., city, state/province, country, and continent. Our strategy performed this alignment by 
first passing the textlabels individually through a pretrained text-encoder and taking the mean of the generated text features resulting in a textual feature vector($F_t$). Then TLA loss($L_{TLA}$) (main paper Eq. \textcolor{red}{3}) is computed between $F_t$ and the output vector from the model ($PV'_t$) obtained by passing the attended features from the self-cross attention module ($PV'$) through an MLP ($FFN_t$). This helped to further improve the location prediction performance of our model. Table \textcolor{red}{3} in the main paper quantitatively demonstrates the same.

In the main paper, we used 
CLIP\cite{radford2021learning} Text Encoder for computing $F_t$ as mentioned. 
Here, we perform an ablation to compare the performance of our method across different choices of pretrained text encoders for computing $F_t$. The results are reported in Table \ref{tab:tla}. 
We observe that the inclusion of the TextLabel Alignment Strategy improves the performance of our model regardless of the text encoder used, as all variants with TLA are better than the one without TLA. Out of all the encoders, features obtained from CLIP(Text) yield the best results. Hence, CLIP(Text) was used for computing $F_t$ in all other experiments. 

\begin{table}[!h]
\centering
\vspace{-0.1in}
\resizebox{0.7\textwidth}{!}{
\begin{tabular}{lcccc}
\hline
\textbf{Text Encoder}         & \multicolumn{1}{l}{\textbf{City}} & \multicolumn{1}{l}{\textbf{State}} & \multicolumn{1}{l}{\textbf{Country}} & \multicolumn{1}{l}{\textbf{Continent}} \\ \hline
w/o TLA      & 67.9                              & 68.4                                        & 72.4                               & 81.6   
\\
OPT\cite{zhang2022opt}                & 68.4                              & 68.9                                        & 72.9                                & 82.3                                  \\
BERT\cite{devlin2019bert}                & 68.7                              & 69.1                                        & 73.8                                & 83.2                                 \\
GPT-2\cite{radford2019language}                & 68.9                              & 69.4                                        & 74.1                                & 83.6                                  \\
\textbf{CLIP(Text)\cite{radford2021learning}} & \textbf{69.6}                    & \textbf{70.2}                              & \textbf{74.8}                       & \textbf{83.8}                         \\ \hline
\end{tabular}
}
\vspace{0.05in}
\caption{\textbf{Ablation Study on Text Encoders} Comparing variants of the model trained with $F_t$ generated from different text encoders. We show that inclusion of TLA strategy works regardless of the choice of text encoder.}
\vspace{-0.6in}
\label{tab:tla}
\end{table}

\section{Top5 Analysis}
\label{sec:top5}
\vspace{-0.1in}
Table \textcolor{red}{5} in the main paper showed the performance of our model in terms of Top1 accuracy at all hierarchies. By definition, top1 accuracy is the measure of how often the most confident prediction of the model is the correct prediction. Similarly, top5 accuracy tells us how often the correct prediction is among the 5 most confident predictions of the model. Even though it is more relaxed, top5 accuracy offers more insights into the performance of a model.

\begin{figure}[!h]
  \centering
  \includegraphics[width = \textwidth]{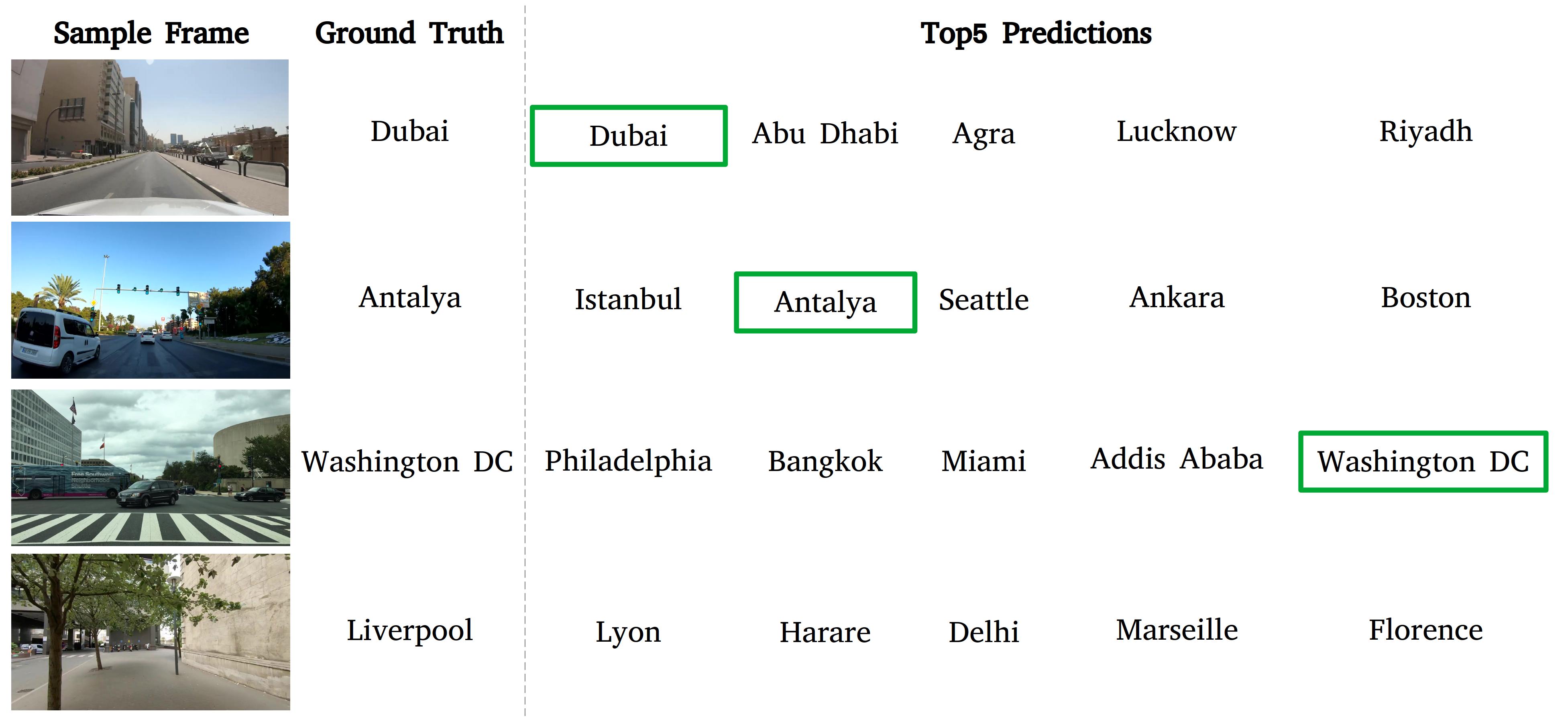}
   \caption{\textbf{Analysis of top5 predictions of a few samples.} First row : Most confident prediction is the correct one; Second and third rows : Most confident prediction is not correct, but correct prediction is among the top5; Last row : Correct prediction is not in the top5. In all cases, most predictions in the top5 are geographically close to the ground truth city}
   \label{fig:top5}
   \vspace{-0.2in}
\end{figure}

Table \ref{tab:top5} shows the top1 and top5 accuracies of our model, at each hierarchy. Naturally, top5 accuracy is a lot higher in each case. Fig. \ref{fig:top5} shows the top5 predictions of a few samples. In the first row, we see that the sample video is correctly classified into Dubai at city-level. Also, the other predictions at top5 contain two cities from the middle-east, which are in close proximity to the ground truth. Rows 2 and 3 show examples where the correct prediction isn't the most confident one, but is among the top5. Again, most of the other predictions in the top5 are close to the ground truth. The last row shows an example where the ground truth is not present in the top 5. Even then, 3 of the 5 predictions are still European cities, which share some traits with Liverpool, where the video is actually from. These observations imply that our model learns reasonable geographical context from a given video.
\begin{table}[htp]
\centering
\vspace{-0.1in}
\resizebox{0.6\textwidth}{!}{
\begin{tabular}{lcc}
\hline
\textbf{Hierarchy} & \multicolumn{1}{l}{\textbf{Top1 Accuracy}} & \multicolumn{1}{l}{\textbf{Top5 Accuracy}} \\ \hline
City               & 69.6                                      & 85.4                                      \\
State     & 70.2                                      & 85.4                                      \\
Country            & 74.8                                      & 88.3                                      \\
Continent          & 83.8                                      & 99.5                                      \\ \hline
\end{tabular}
}
\vspace{0.05in}
\caption{\textbf{Top5 Analysis.} Performance of our model in terms of Top1 and Top5 accuracies for additional insights}
\vspace{-0.2in}
\label{tab:top5}
\end{table}

\section{Data preparation for Mapillary(MSLS)}
\label{sec:data-prep}
Mapillary(MSLS)\cite{warburg2020mapillary} dataset in total has $24$ cities in its train set, and $6$ cities in its test set. Also we do not train and validate on different cities, as the problem requires all classes to be represented in both train and validation sets. 
We combined the videos from train and test sets and redistributed them according to an $80$:$20$ stratified split, such that all classes are fairly represented in both sets. 

Once, the entire combined list of sequences was collected, filtering was necessary to make the dataset compatible. As specified in Section \textcolor{red}{5.2} in the main paper, the model inputs videos that are $15$ frames in length. Thus we had to remove all sequences which had less than $15$ frames. Also to be consistent with our approach till now, we had to take sequences from only those cities, that were present in our dataset. After filtering these sequences, we were finally left with $11,320$ samples from $21$ cities, $21$ states/provinces, $19$ countries and $6$ continents. For soft-scene labels, we followed a similar procedure (Main paper Section \textcolor{red}{4.3}) as for our dataset.
\section{Post processing procedures for Anonymity preservation}
\label{sec:x}
\begin{figure}[!t]
  \centering
  \includegraphics[width = \textwidth]{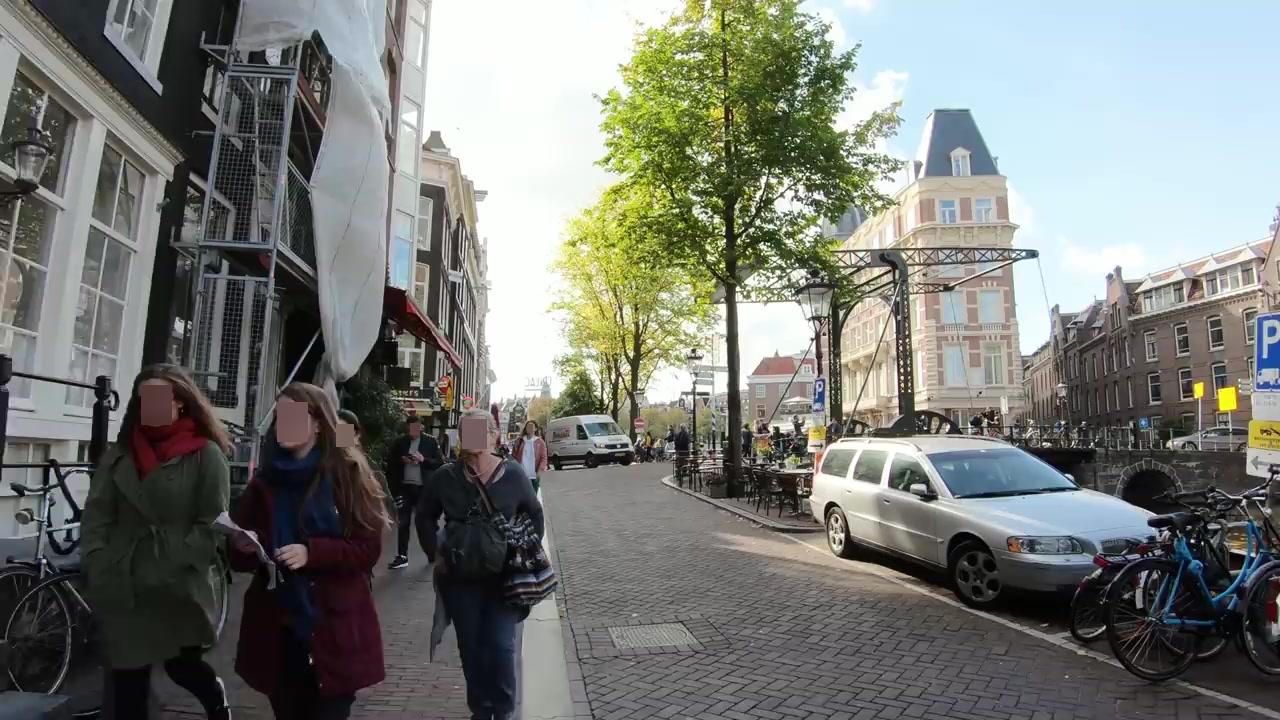}
  \caption{\textbf{Sample frame with blurred faces.} Location: Amsterdam, Netherlands}
  \label{fig:blur}
  \vspace{-0.2in}
\end{figure}

As highlighted in Section \textcolor{red}{3} in the main paper, we make an extensive effort for preserving Anonymity of individuals that may have appeared in some videos of our dataset. The procedure we followed is as follows:
\begin{enumerate}
    \item \textbf{Face Detection:} First, we scanned all the frames of all videos for faces. For this purpose we used RetinaFace\cite{deng2020retinaface}. For the detector, the confidence threshold was set to 0.5 and the NMS\_IoU threshold was set to 0.3
    \item \textbf{Manual Inspection:} The first frame with detected faces from each video was inspected manually to filter images with issues like false positives, far away faces, unclear faces, etc 
    \item \textbf{Face blurring:} The detected faces from each frame were blurred out one by one using the GaussianBlur function from Python's opencv library. The kernel size was set to 63 and the standard deviation was set to 21 for the function. Fig. \ref{fig:blur} shows a sample frame from a video taken in Amsterdam, Netherlands with a high face count. We observe that all the faces present in the frame have been blurred correctly.
\end{enumerate}

\section{Analysis of CityGuessr68k Dataset}
\label{sec:cg68-analysis}
In this section, we discuss and analyze our proposed dataset `CityGuessr68k' (Section \textcolor{red}{3} in the main paper). For a closer insight into our proposed dataset, we provide more visualizations in terms of sample video frames and distribution analysis curves. We also compare our dataset with the Mapillary(MSLS)\cite{warburg2020mapillary} image sequence dataset using various metrics.
\vspace{-0.1in}
\subsection{Additional Visualizations of Data Samples}
\label{sec:viz}
CityGuessr68k dataset comprises of $68,269$ videos from $166$ cities, $157$ states/ provinces, $91$ countries and $6$ continents spread all over the world. To provide a quick peek into our dataset, we showcase extensive visualizations in the form of sample frames representing one video from every city. The visualizations are arranged continent-wise for convenience. Fig. \ref{fig:data-viz} showcases the sample video frames representing cities from countries in Africa(Fig. \ref{fig:africa}), Europe(Fig. \ref{fig:europe}), North America(Fig. \ref{fig:northamerica}), South America(Fig. \ref{fig:southamerica}), Asia(Fig. \ref{fig:asia}) and Oceania(Fig. \ref{fig:oceania}) in that order. 

\begin{figure}[p]

  \centering
  \begin{subfigure}{\linewidth}
    \centering
    \includegraphics[width = \textwidth]{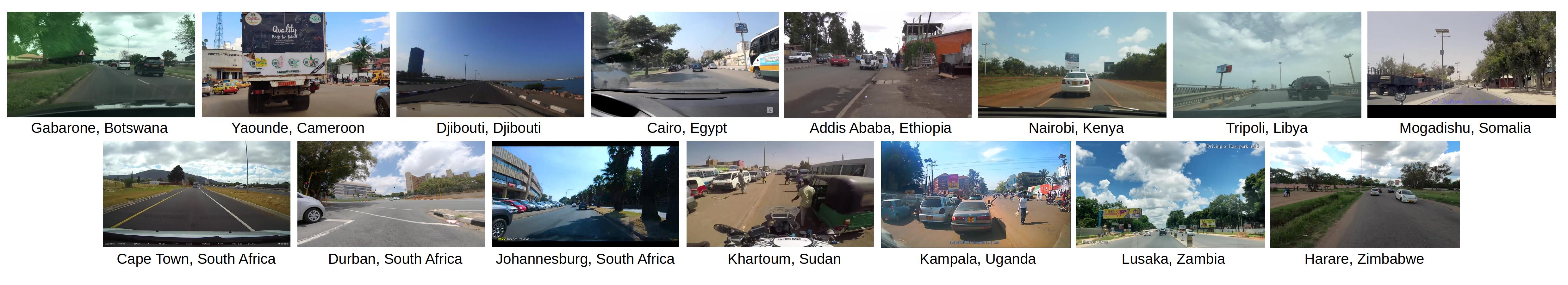}
    \caption{Visualizations of sample video frames from each city in \textbf{Africa}}
    \label{fig:africa}
  \end{subfigure}
  \begin{subfigure}{\linewidth}
    \centering
    \includegraphics[width = \textwidth]{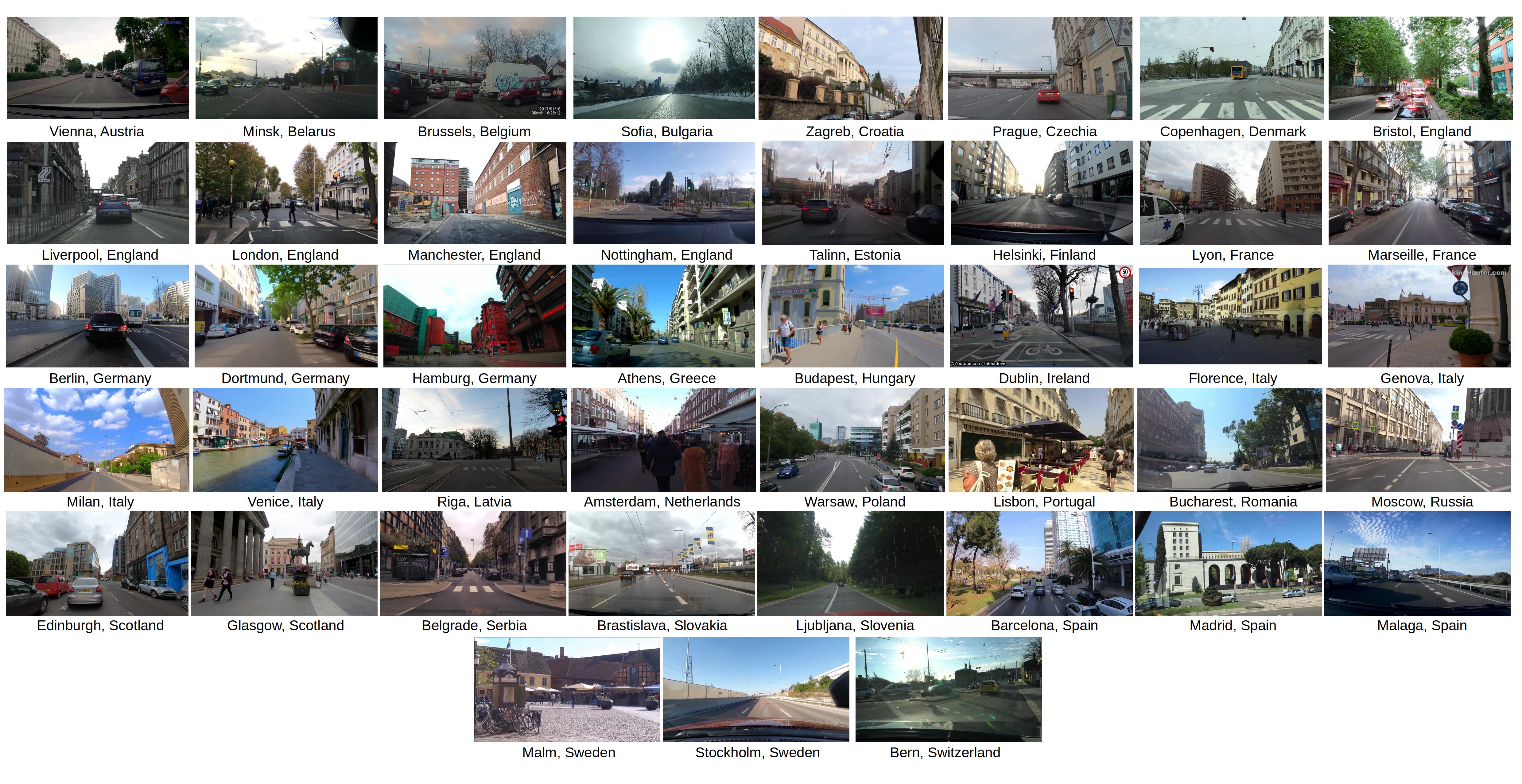}
    \caption{Visualizations of sample video frames from each city in \textbf{Europe}}
    \label{fig:europe}
    
  \end{subfigure}
  \medskip
  \begin{subfigure}{\linewidth}
    \centering
    \includegraphics[width = \textwidth]{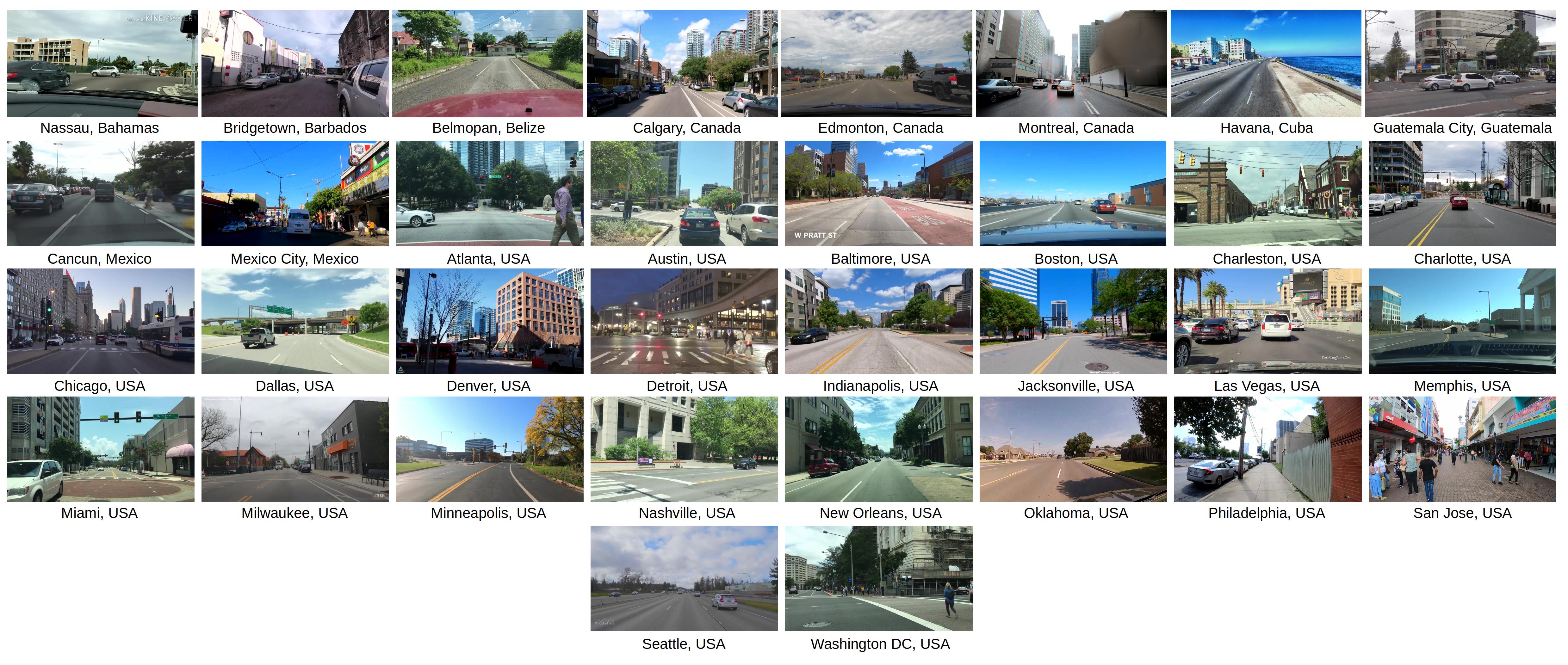}
    \caption{Visualizations of sample video frames from each city in \textbf{North America}}
    \label{fig:northamerica}
  \end{subfigure}
  \begin{subfigure}{\linewidth}
    \centering
    \includegraphics[width = \textwidth]{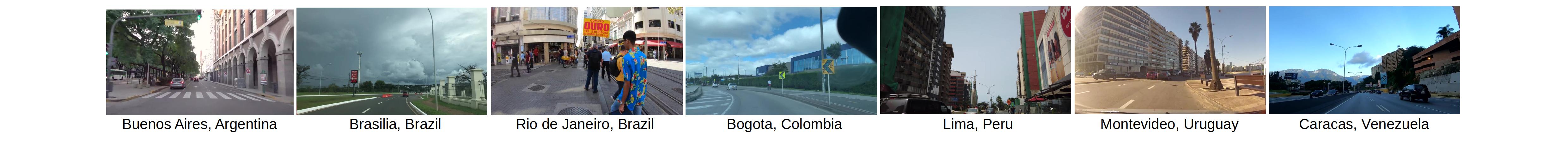}
    \caption{Visualizations of sample video frames from each city in \textbf{South America}}
    \label{fig:southamerica}
  \end{subfigure}
\end{figure}
\begin{figure} \ContinuedFloat
  \centering
  \begin{subfigure}{\linewidth}
    \centering
    \includegraphics[width = \textwidth]{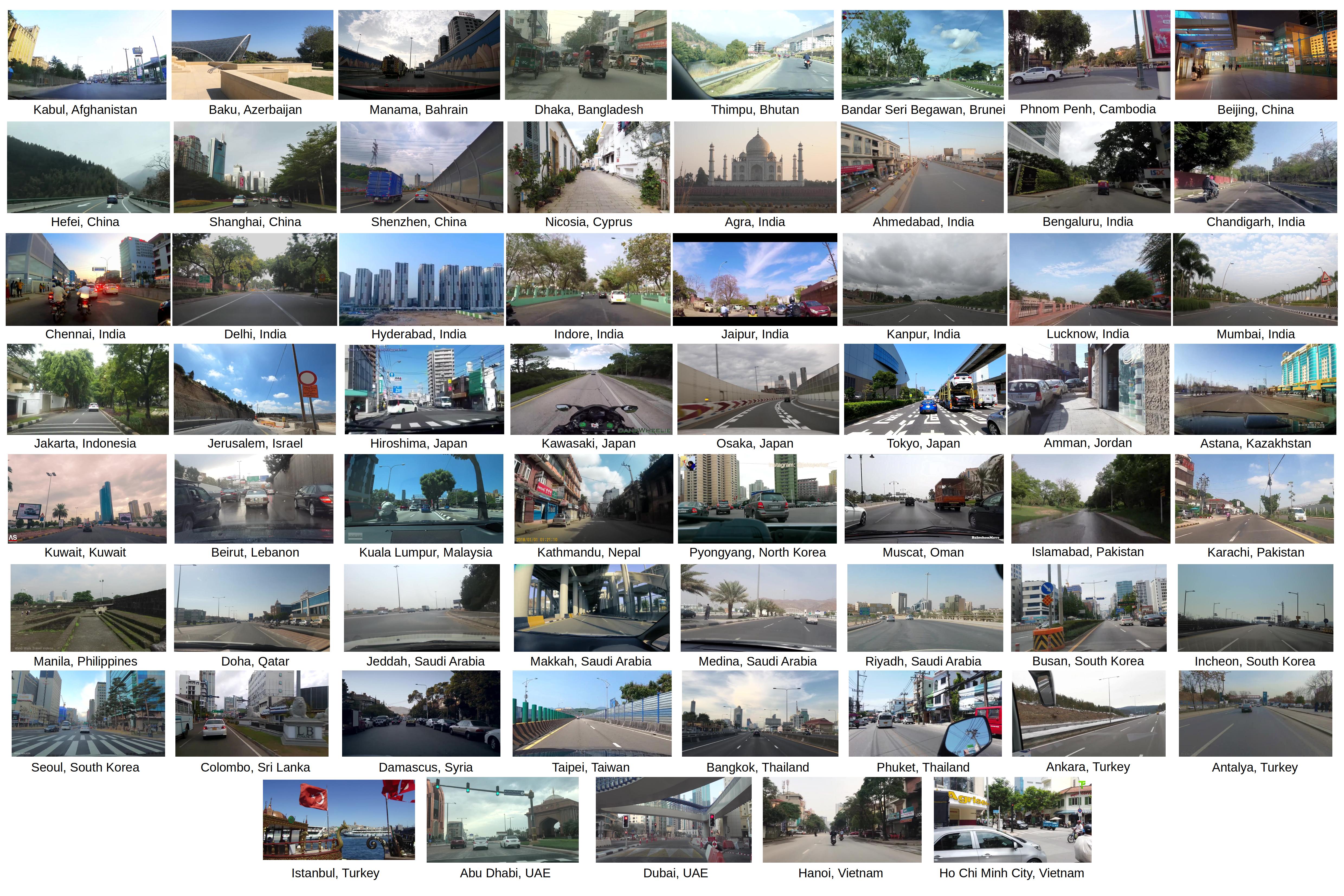}
    \caption{Visualizations of sample video frames from each city in \textbf{Asia}}
    \label{fig:asia}
  \end{subfigure}
  \begin{subfigure}{\linewidth}
    \centering
    \includegraphics[width = \textwidth]{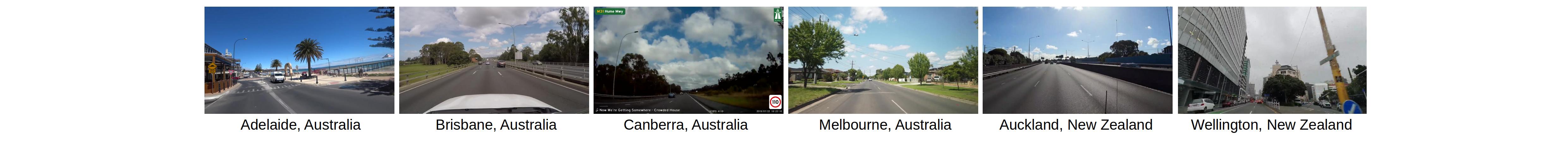}
    \caption{Visualizations of sample video frames from each city in \textbf{Oceania}}
    \label{fig:oceania}
  \end{subfigure}
   
   \caption{\textbf{Extensive Visualizations of sample videos from all cities in CityGuessr68k.} This figure shows one frame from a sample video from every city in CityGuessr68k. The dataset contains videos from 166 cities, 157 states/provinces, 91 countries and 6 continents. The figure is divided continent-wise for better insights. In order, Africa has 15 cities, Europe has 43, North America has 34, South America 7, Asia 61 and Oceania has 6. As we can see the dataset is very diverse in terms of scenes, infrastructure and activities(driving, walking, boating, biking, etc.)}
   \label{fig:data-viz}
\end{figure}
\newpage
\subsection{Lorenz Curves}
\label{sec:lorenz}
\begin{figure}[h]
  \centering
  \includegraphics[width = 0.7\textwidth]{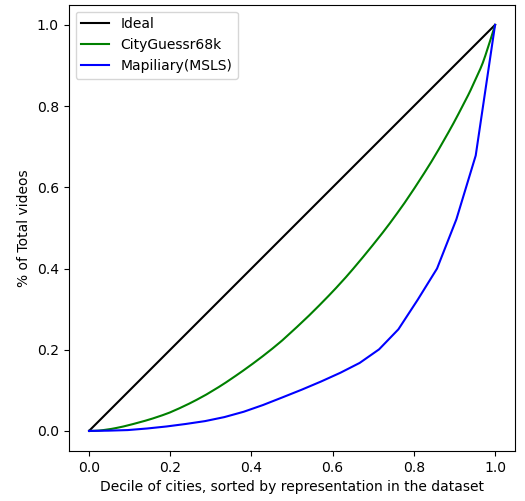}
   \caption{\textbf{Comparison of Lorenz curves between CityGuessr68k and Mapillary(MSLS) datasets.} $x$-axis represents the cities, sorted by cumulative number of samples, which are represented by the $y$-axis. Black line is the Lorenz curve of an ideal distribution. CityGuessr68k's curve is closer to the ideal line, implying a more even distribution.}
   \label{fig:lorenz}
\end{figure}

Lorenz curve\cite{lorenz1905methods} is a graphical representation of the distribution of data (originally, of income or of wealth). The concept was initially formulated to represent inequality of wealth distribution. Eventually, it found its way into studying inequality of distributions of non-economic data as well. We can use this concept to demonstrate the global evenness of distribution of samples in our dataset. We plot the Lorenz curves for both CityGuessr68k and Mapillary(MSLS)\cite{warburg2020mapillary} in Fig. \ref{fig:lorenz}. We also plot the curve for a perfectly even distribution for reference. A Lorenz curve starts at $(0,0)$ and ends at $(1,1)$. The $x$-axis represents the cities, sorted by the cumulative number of samples. The $y$-axis represents the cumulative number of samples up to a particular class. The ideal Lorenz curve is the $y = x$ line which implies a perfect distribution, as the cumulative number of samples increases linearly with the number of classes. We observe that, there is inequality present in both datasets, but CityGuessr68k has a much more even distribution as it is closer to the ideal curve than Mapillary(MSLS). The Lorenz curves demonstrate that videos are spread more evenly throughout the cities in the world in the case of CityGuessr68k, while Mapillary(MSLS) seems to be more biased(skewed towards certain classes).

\subsection{Gini Coefficient}
\label{sec:gini}
Gini coefficient\cite{gates1909mutability} is another measure of inequality within a distribution, which also comes from economics. Mathematically, Gini coefficient is the ratio of the area that lies between the line of equality (perfectly even distribution) and the Lorenz curve, over the total area under the line of equality. Thus, a Gini coefficient of $0$ implies perfect equality while a Gini coefficient of $1$ means maximal inequality, a situation where a single class has all the samples while all others have none. Let $y_i$ be the number of samples in the $i^{th}$ class, in a dataset with $n$ classes. Then, the Gini coefficient is defined as 
\begin{align}
    G = \frac{2\sum_{i=1}^{n}iy_i}{n\sum_{i=1}^{n}y_i} - \frac{n+1}{n}
\end{align}
We calculate Gini coefficients for both datasets. We find CityGuessr68k's Gini coefficient to be $0.36$, while the Gini coefficient of Mapillary(MSLS) is $0.65$. This implies that CityGuessr68k has a much more even distribution as compared to Mapillary(MSLS), considering the big improvement of $0.29$. Thus, Gini coefficient confirms our observations with the Lorenz curve and gives a quantitative measure for the same. 

\subsection{Hoover Index}
\label{sec:hoover}
Hoover index\cite{hoover1941interstate} is the measure of inequality (also from economics), in terms of redistribution of data. It is equal to the percentage of the number of samples that would have to be redistributed to make the number of samples in all classes equal. Mathematically, Hoover index is the longest vertical distance between the Lorenz curve and the line of equality (perfectly even distribution). Thus a Hoover index of $0$ implies perfect equality, as there is no requirement to redistribute any data, while a Hoover index of $1$ means maximal inequality, where the entire data needs to be redistributed for even distribution. Let $y_i$ be the number of samples in the $i^{th}$ class, in a dataset with $n$ classes. Then, the Hoover index is defined as 
\begin{align}
    H = \frac{1}{2}\frac{\sum_{i=1}^{n}\left | y_i - \bar{y} \right |}{\sum_{i=1}^{n}y_i}
\end{align}
where $\bar{y}$ is the mean number of samples in a class. We calculate Hoover Index for CityGuessr68k and Mapillary(MSLS). CityGuessr68k's Hoover index is $0.26$ while Hoover index for Mapillary(MSLS) is $0.51$, which implies that Mapillary will require a lot more redistribution compared to CityGuessr68k to achieve a perfectly even distribution. This once again affirms our previous findings with other metrics, that CityGuessr68k has a significantly more even distribution of samples than Mapillary(MSLS).
\newpage
\section{Qualitative Analysis}
\label{sec:qual}
In this section, we perform a qualitative analysis of the localization results predicted by our model, to support the quantitative results showcased in Section \textcolor{red}{5} in the main paper. In Section \ref{sec:correct}, we discuss the distribution of correct predictions on the world map, and Section \ref{sec:loc-eg} provides localization examples at different hierarchies.
\begin{figure}[h]
    \centering
    \includegraphics[width = 0.55\textwidth]{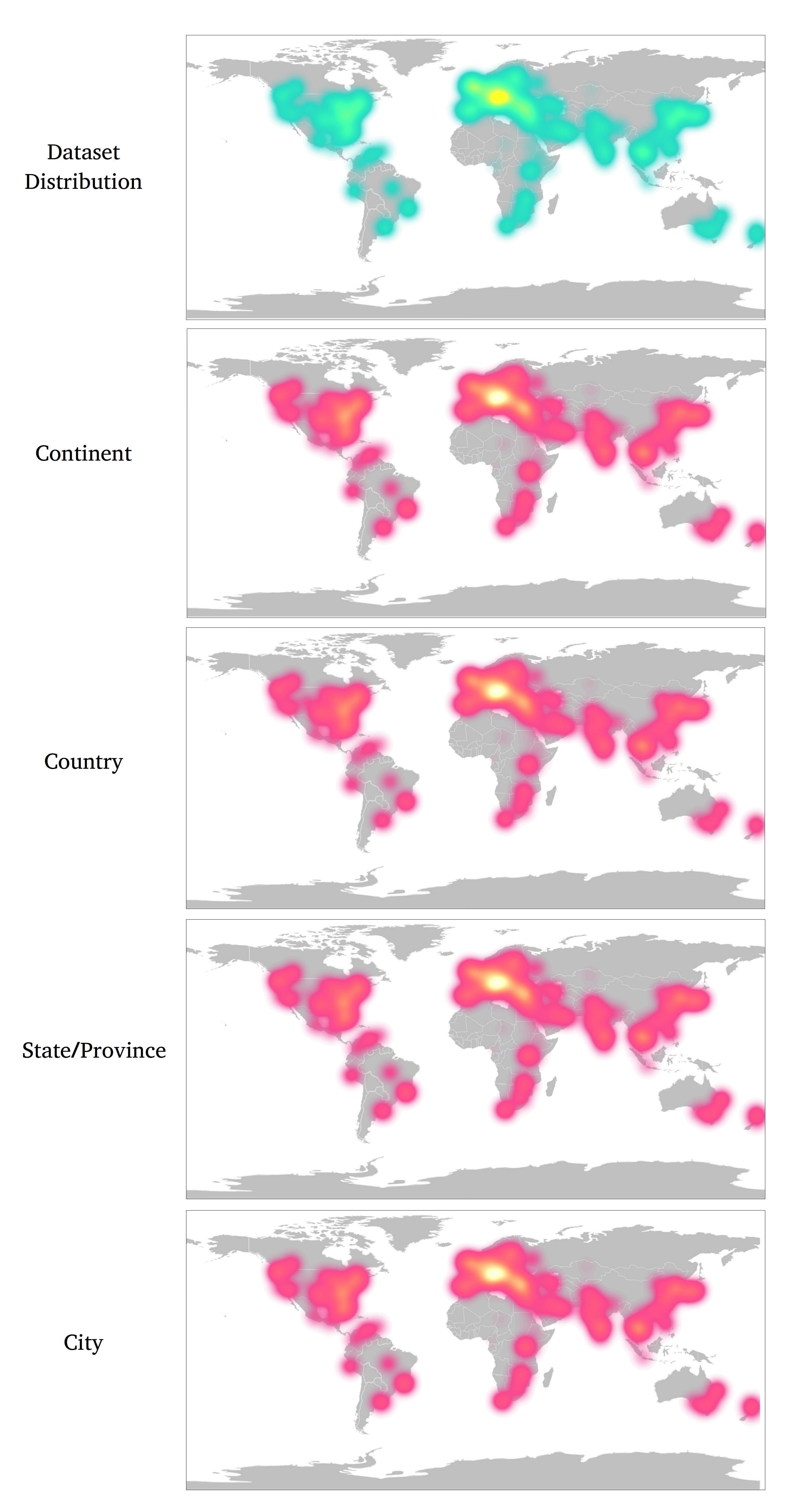}
    \caption{\textbf{A visualization of the distribution of the model's correct predictions for different hierarchies on a world map.} Similarity in the distributions implies successful location prediction down to the finest hierarchy. Intensity decreases for finer hierarchies as accuracy drops}
    \label{fig:pred-dist}
\end{figure}
\newpage
\subsection{Correct Predictions Around the World}
\label{sec:correct}
Fig. \ref{fig:pred-dist} shows the distribution of the locations correctly predicted by our model for each hierarchy, on the world map. The overall distribution of the dataset is shown at the top for reference. We observe that the distributions for all 4 hierarchies are very similar, only differing in intensity. This is an indicator that our model is able to successfully predict locations evenly around the globe down to the finest hierarchy, and is not biased towards any specific region. The decreasing intensity is due to the drop in accuracy as we go towards finer hierarchies.

\subsection{Examples of Localizations}
\label{sec:loc-eg}
\begin{figure*}[h]
    \centering
    \includegraphics[width=\textwidth]{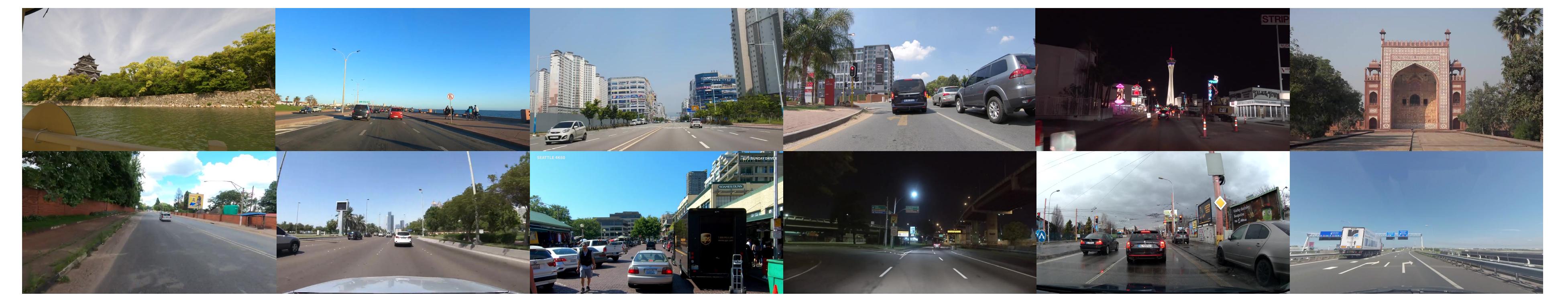}
    \caption{\textbf{Visualization of random samples of the dataset for which our model correctly predicted the city.} The features of these video frames mostly include landmarks or highlight infrastructure that can be specifically found in that city. Sign boards are also visible which might be helpful to the model for accurate prediction at the finest hierarchy.}
    \label{fig:city-loc}
\end{figure*}

\begin{figure*}[h]
    \centering
    \includegraphics[width=\textwidth]{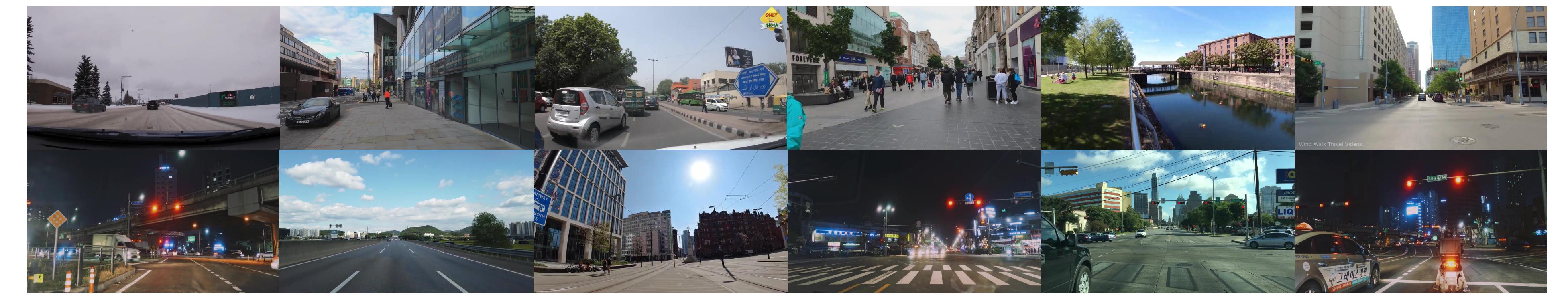}
    \caption{\textbf{Visualization of random samples of the dataset for which our model could not predict the city correctly, but predicted the video to be in the same state/province as the correct prediction.} The features of these video frames include a lot of things that are native to a particular region. That includes certain non-urban locations as well along with road and weather patterns.}
    \label{fig:state-loc}
\end{figure*}

\begin{figure*}[h]
    \centering
    \includegraphics[width=\textwidth]{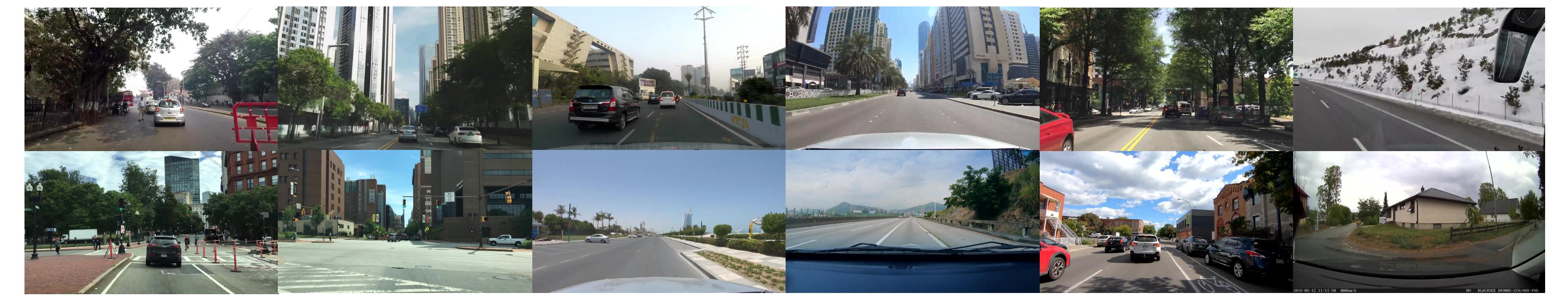}
    \caption{\textbf{Visualization of random samples of the dataset which were predicted to be in the correct country, but were mispredicted into the wrong state/province.} The videos have become more challenging. The features that the model can rely on are mostly geographical. Infrastructure-wise if there are specific style of buildings native to a country, that might help. Even the side of the road the vehicle is driving on, is helpful.}
    \label{fig:country-loc}
\end{figure*}

\begin{figure*}[h]
    \centering
    \includegraphics[width=\textwidth]{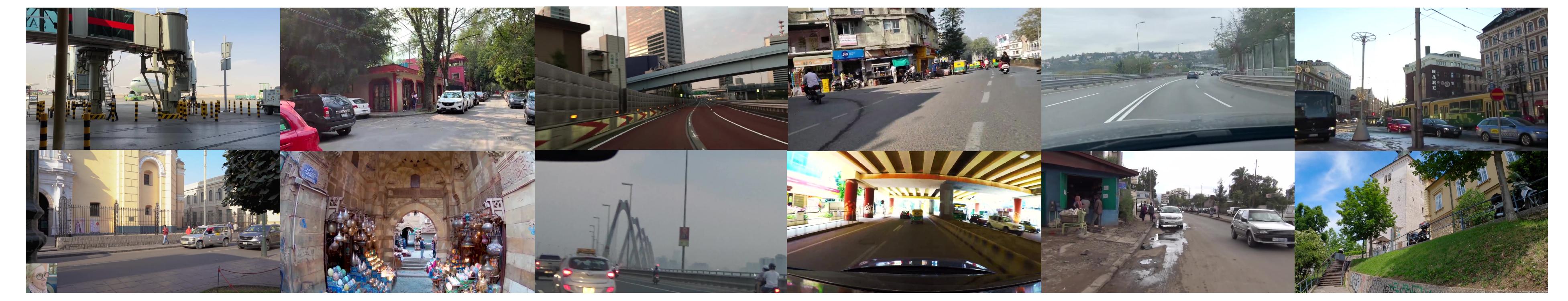}
    \caption{\textbf{Visualization of random samples of the dataset which were correctly predicted within the continent, but the finer hierarchies were predicted erroneously.} The features in the video are primarily cultural, which might help in identifying a coarse area. Specific infrastructure is helpful, but with a high uncertainty.}
    \label{fig:continent-loc}
\end{figure*}

\begin{figure*}[h]
    \centering
    \includegraphics[width=\textwidth]{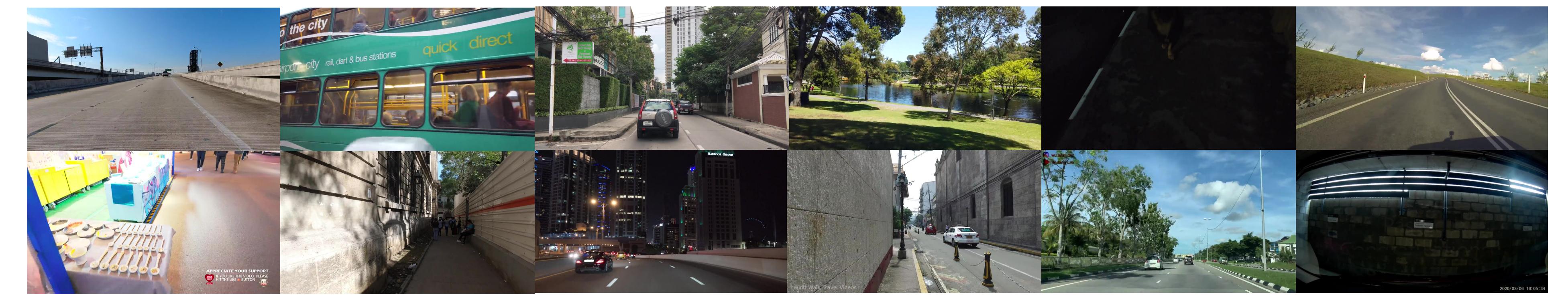}
    \caption{\textbf{Visualization of random samples of the dataset  that were predicted incorrectly in all hierarchies, where the model failed to recognize any relevant features.} The features in the video are very general and could be found anywhere in the world.}
    \label{fig:fail-loc}
\end{figure*}

The model focuses on different attributes in a video to localize within every hierarchy. The features in the video that the model might use for identifying the city, might be very different from the features it uses for identifying the state/province or country or continent. This is apparent, as the identification of city will utilize finer features concentrated in a very specific part of the world, while the identification of a continent requires more general features which are seen over a vast expanse of land. We demonstrate this by showing sample frames from successful predictions up to every hierarchy. Fig. \ref{fig:city-loc} shows frames from a few videos for which the city and consequently all coarser hierarchies were correctly predicted. Fig. \ref{fig:state-loc} shows the samples where the model could not predict the city correctly, but predicted the video to be in the same state/province as the correct prediction. Fig. \ref{fig:country-loc} shows samples that were predicted to be in the correct country, but were mispredicted into the wrong state/province (and city by association). Fig. \ref{fig:continent-loc} shows a few examples of videos that were correctly predicted within the continent, but the finer hierarchies were predicted erroneously. Finally Fig. \ref{fig:fail-loc} shows some videos that were predicted incorrectly in all hierarchies, where the model failed to recognize any relevant features.



\clearpage

\end{document}